\documentclass{article} 
\usepackage{collas2024_conference,times}
\usepackage{easyReview}

\usepackage{amsmath,amsfonts,bm}

\def\eqref#1{equation~\ref{#1}}

\def\1{\bm{1}}

\DeclareMathAlphabet{\mathsfit}{\encodingdefault}{\sfdefault}{m}{sl}
\SetMathAlphabet{\mathsfit}{bold}{\encodingdefault}{\sfdefault}{bx}{n}

\usepackage{hyperref}
\hypersetup{
    colorlinks=true,
    linkcolor=red,
    filecolor=magenta,
    urlcolor=blue,
    citecolor=purple,
    pdftitle={Generative Distillation},
    pdfpagemode=FullScreen,
    }

\usepackage[utf8]{inputenc} 
\usepackage[T1]{fontenc}    
\usepackage{url}            
\usepackage{booktabs}       
\usepackage{amsfonts}       
\usepackage{amssymb}
\usepackage{algorithm}
\usepackage{algpseudocode}
\usepackage{nicefrac}       
\usepackage{microtype}      
\usepackage{xcolor}         
\usepackage{graphicx}
\usepackage{subcaption}
\usepackage{mathtools}
\usepackage{cleveref}
\usepackage{siunitx}
\usepackage{svg}
\usepackage{wrapfig}
\usepackage{enumitem}
\usepackage{adjustbox}
\usepackage{enumitem}

\newcommand{\Exp}{\mathop{\mathbb{E}}}
\newcommand{\Teacher}{\epsilon_{\hat{\theta}_{i-1}}}
\newcommand{\Student}{\epsilon_{\theta_{i}}}

\title{Continual Learning of Diffusion Models with Generative Distillation}

\author{%
  \hspace{0.25in}Sergi Masip\thanks{Work done under the Erasmus+ Traineeship program at KU Leuven.} \\
  \hspace{0.25in}Master in Computer Vision (Barcelona)\\
  \hspace{0.25in}\texttt{hello@sergimasip.com} 
  \And
  Pau Rodríguez \\
  Apple\thanks{Work done externally as part of the Master in Computer Vision (Barcelona).}\\
  \texttt{pau.rodriguez@apple.com}\hspace{0.72in}
  \And
  \hspace{0.25in}Tinne Tuytelaars \\
  \hspace{0.25in}KU Leuven \\
  \hspace{0.25in}\texttt{tinne.tuytelaars@esat.kuleuven.be}
  \And
  Gido M. van de Ven \\
  KU Leuven \\
  \texttt{gido.vandeven@kuleuven.be} 
}

\collasfinalcopy 

\begin{document}

\maketitle

\begin{abstract}
Diffusion models are powerful generative models that achieve state-of-the-art performance in image synthesis. However, training them demands substantial amounts of data and computational resources. Continual learning would allow for incrementally learning new tasks and accumulating knowledge, thus enabling the reuse of trained models for further learning. One potentially suitable continual learning approach is generative replay, where a copy of a generative model trained on previous tasks produces synthetic data that are interleaved with data from the current task. However, standard generative replay applied to diffusion models results in a catastrophic loss in denoising capabilities. In this paper, we propose \textit{generative distillation}, an approach that distils the entire reverse process of a diffusion model. We demonstrate that our approach substantially improves the continual learning performance of generative replay with only a modest increase in the computational costs.
\end{abstract}

\section{Introduction}
Diffusion models for likelihood-based data generation \citep{song2020generative, ho2020denoising, song2021scorebased} have shown impressive results in tasks such as conditional and unconditional image generation. 
However, a disadvantage of these models is that they must be trained on vast datasets while using large amounts of computational resources, making their training very costly. Furthermore, once a diffusion model has been trained, incorporating new knowledge incrementally is challenging as the performance on previously learned tasks severely decreases when trained on new tasks. This phenomenon is known as catastrophic forgetting~\citep{MCCLOSKEY1989109, Ratcliff1990ConnectionistMO} and it is common when training neural networks in a sequential manner~\citep{survey_cl}. For this reason, deep learning practitioners typically retrain their models using all the data seen so far each time they incorporate new data. Especially for diffusion models, this is an extremely costly process. It would therefore be valuable to devise a way to leverage already trained diffusion models more efficiently and allow them to continuously expand their knowledge.

The continual learning field studies how to mitigate catastrophic forgetting to enable neural networks to incrementally learn new tasks, domains or classes while retaining previous knowledge \citep{van2022three}. Among the existing approaches for continual learning, rehearsal methods, which store a fixed replay buffer of samples from previous tasks, have shown promising initial results for diffusion models \citep{zajac2023exploring}. However, previous data are sometimes unavailable or their use might be undesirable or even infeasible (e.g., due to privacy policies or computational cost). An alternative is generative replay~\citep{shin2017}, where the replay buffer is replaced with a generative model that produces synthetic samples. Previous works \citep{zajac2023exploring, gao2023ddgr} that used generative replay for continually training diffusion models employed the original sampler from ``Denoising Diffusion Probabilistic Models'' (DDPM) \citep{ho2020denoising} to generate the synthetic samples. 

An important limitation of DDPM is that its generation times are prohibitively slow. This restriction forced \citet{zajac2023exploring} and \citet{gao2023ddgr} to generate a fixed buffer of synthetic samples, thereby limiting the diversity advantage that diffusion models could offer. An alternative solution is to employ the sampler introduced in ``Denoising Implicit Probabilistic Models'' (DDIM)~\citep{song2020denoising}, which allows for much faster generation in exchange for a small reduction in image quality, enabling generative replay to use a considerably larger amount of samples. Nonetheless, we find that generative replay with the approximate samples yielded by DDIM results in a fast decrease in image quality, which seems to be due to a catastrophic loss in the model's ability to denoise images from previous tasks. 
Previous works have observed severe forgetting with generative replay for diffusion models also when using DDPM \citep{smith2023continual,gao2023ddgr}.
We hypothesize that the reason for this is that, in the case of diffusion models, the transfer of knowledge in standard generative replay is rather inefficient.
As illustrated in \Cref{fig:gen_replay}, when training on a new task, standard generative replay transfers knowledge from the previous task model to the current task model only at the endpoint of the reverse process of the diffusion model (i.e., through the generated replay samples). 

To improve generative replay for diffusion models, in this paper we propose to transfer knowledge more effectively by combining generative replay with knowledge distillation. This approach, which we call \textit{generative distillation}, distils the entire reverse process of a \textit{teacher} diffusion model trained on previous tasks to the \textit{student} model being trained on the current task. The teacher generates a sample with DDIM, then Gaussian noise is added according to the variance schedule~\citep{ho2020denoising} and, finally, the objective matches the noise predictions of both the teacher and the student (\Cref{fig:gen_distil}). We conduct experiments on Fashion-MNIST~\citep{xiao2017fashion} and CIFAR-10~\citep{krizhevsky2009learning} where we continually train a diffusion model with generative distillation and demonstrate its effectiveness compared to generative replay at mitigating forgetting and the catastrophic loss in denoising capabilities, while increasing the computational cost by just one forward pass.

The main contributions of this paper can be summarized as follows:

\begin{itemize} 
  \item We show that continually training diffusion models using standard generative replay with DDIM results in a catastrophic loss in denoising capabilities (\Cref{sec:catastrophic}).
  \item To mitigate this issue, we introduce generative distillation, an approach that combines generative replay with knowledge distillation (\Cref{sec:methods-generative-distillation}).
  \item We demonstrate that generative distillation mitigates the loss in denoising capabilities and substantially improves upon generative replay, while using almost the same computational budget (\Cref{sec:exps-generative-distillation}).
  \item We further highlight the effectiveness of distillation for the continual training of diffusion models in additional experiments by applying it to input images of various quality, including Gaussian noise (\Cref{experiments:extra}).
\end{itemize}

\section{Background and related work}

This section provides a brief description of methods and literature on diffusion models, distillation, and continual learning that are relevant to our work.

\subsection{Diffusion models} 
Diffusion models are likelihood-based models~\citep{song2020generative, ho2020denoising, song2021scorebased} defined by a forward process that adds noise to an image and a reverse process that denoises it. 
The forward process is a Markov chain with transitions that are Gaussian conditional distributions as defined by:
\begin{equation}\label{eq:forward_process}
    q(x_{1:T}|x_0)=\prod_{t=1}^T q(x_t|x_{t-1}),\qquad q(x_t|x_{t-1})=\mathcal{N}(x_t; \sqrt{1-\beta_t}x_{t-1}, \beta_tI).
\end{equation}
This Markov chain progressively adds noise to an image $x_0$ through $x_1, x_2, ..., x_T$ according to a variance schedule~$\beta_t$.
By setting $\alpha_t := 1-\beta_t$ and $\bar{\alpha}:=\prod_{s=1}^t \alpha_s$ so that $(1-\bar{\alpha}_t)$ expresses the variance of the Gaussian noise, the above process can be reparameterized as:
\begin{equation}\label{eq:reparametrization_forward}
q(x_t|x_0)=\mathcal{N}(x_t; \sqrt{\bar{\alpha}_t} x_0, (1-\bar{\alpha}_t)I),
\end{equation}
allowing for sampling $x_t\sim q(x_t|x_0)$ at any timestep $t$ in just one step. 

The reverse process follows the reverse trajectory of the forward process. 
Although, in general, the shape of the posterior $q(x_{t-1}|x_t)$ is unknown, when $\beta_t \rightarrow 0$, it converges to a Gaussian \citep{sohl2015deep}. Consequently, $q(x_{t-1}|x_t)$ can be approximated by modelling the mean and the variance of a conditional Gaussian distribution through a variational distribution $p_\theta(x_{t-1}|x_t)$. This gives the following reverse process:
\begin{equation}\label{eq:reverse_process}
    p_\theta(x_{0:T}) = p(x_T) \prod_{t=1}^T p_\theta(x_{t-1}|x_t),\qquad p_\theta(x_{t-1}|x_t)=\mathcal{N}(x_{t-1}; \mu_\theta(x_t, t), \Sigma_\theta(x_t, t)).
\end{equation}
with $\theta$ the learnable parameters of the diffusion model.

A diffusion model, which from now on we will denote as $\epsilon_\theta$, can be trained to match the noise introduced during the forward process at any timestep~$t$ following a predefined variance schedule. This can be achieved by minimizing the following simplified objective~\citep{ho2020denoising}: 
\begin{equation}\label{eq:gr_loss_generator_simple}
    L_\text{simple}(\theta) = \mathbb{E}_{t \sim \text{Unif}(T), x_0 \sim D, \epsilon \sim \mathcal{N}(0, I)} \bigr[ 
        \|\epsilon - \epsilon_\theta(\sqrt{\alpha_t}x_0+\sqrt{1-\alpha_t}\epsilon, t)\|^2
    \bigr],
\end{equation}
where $x_0$ is a clean sample from the observed data distribution~$D$, $\epsilon \sim \mathcal{N}(0, I)$ is the ground truth Gaussian noise injected during the forward pass, and $\alpha$ is a variance schedule function. 

The standard way of sampling from a diffusion model (e.g., with DDPM) is computationally expensive as it requires many denoising steps (e.g., 1000 in \citealp{ho2020denoising}). This greatly limits the number of samples that can be generated for replay with a finite computational budget. An alternative way of sampling is provided by DDIM~\citep{song2020denoising}, which makes the forward process non-Markovian by reformulating $q(x_{t-1}|x_t, x_0)$ so that $q(x_t|x_{t-1})$ also depends on $x_0$. DDIM also introduces a stochastic parameter $\sigma$; when this parameter is set to zero, the sampling process becomes deterministic. DDIM permits sampling using a smaller number of denoising steps, thus allowing the generation of more images given the same computational budget. However, the fewer denoising steps are used, the blurrier and less detailed the sampled images tend to be. Each DDIM denoising step from $t \in T$ to $s \in T$ (such that $s < t$) can be expressed as:
\begin{equation}\label{eq:ddim_sampling}
    x_{s} = \sqrt{\alpha_{s}} \biggl(\frac{x_t - \sqrt{1-\alpha_t}\epsilon_\theta(x_t)}{\sqrt{\alpha_t}} \biggr) 
     + \sqrt{1-\alpha_{s}-\sigma_t^2} \cdot \epsilon_\theta(x_t)
     + \sigma_t \epsilon.
 \end{equation}

\subsection{Knowledge distillation in diffusion models} 
Existing studies use knowledge distillation in diffusion models to train a student network such that it can generate the same quality images as a teacher network, but in fewer generation steps~\citep{luhman2021knowledge, salimans2022progressive, berthelot2023tract, song2023consistency, gu2023boot}. In this paper, we use distillation for a different purpose: we demonstrate its effectiveness for continual learning of diffusion models.

\subsection{Continual learning} An important problem in continual learning is catastrophic forgetting \citep{MCCLOSKEY1989109, Ratcliff1990ConnectionistMO}, a phenomenon where a sequentially trained neural network rapidly and drastically forgets its knowledge of previous tasks when it learns a new task. Current approaches that aim to mitigate catastrophic forgetting can be roughly categorized as architectural or parameter isolation methods, regularization methods and rehearsal methods \citep{biesialska-etal-2020-continual, survey_cl}. Architectural methods \citep{rusu2016progressive, yoon2017lifelong, Scardapane_2017, Masse_2018, wortsman2020supermasks, activedendrite} prevent interference by introducing new modules or by using a fixed set of parameters for each task. Regularization methods \citep{Kirkpatrick_2017, synapticintel, li2017learning, ahn2019uncertaintybased} preserve the knowledge from previous tasks through the addition of extra terms to the loss function. Rehearsal methods \citep{Ratcliff1990ConnectionistMO, rehearsal1995, rebuffi2017icarl, lopezpaz2022gradient, chaudhry2019efficient} complement the training data of the current task with data representative of previous tasks.

Rehearsal is often implemented as experience replay, which uses an episodic buffer to store all or a subset of previous tasks' samples and replays them during the training of later tasks. 
In some situations, there may be privacy issues or compute constraints that forbid storing or using previous data. A popular approach to mitigate catastrophic forgetting under these circumstances is generative replay \citep{shin2017}. Given a sequence of $B$ tasks $\mathcal{T}=\{\mathcal{T}_1, \mathcal{T}_2, ..., \mathcal{T}_B\}$, a generative model~$G_i$ is trained on data from task~$T_i$ interleaved with replay samples produced by $G_{i-1}$, a copy of the generative model after it finished training on the previous task.


\subsection{Continual learning and diffusion models} The topic of continual learning of diffusion models remains relatively unexplored. We are aware of three works that use generative replay in diffusion models. \citet{zajac2023exploring} test different continual learning approaches on diffusion models, and \citet{gao2023ddgr} use a continually trained classifier to guide the generation process, as done by \citet{dhariwal2021diffusion}. These works, however, use the DDPM sampler and do not study how to improve generative replay for diffusion models through the addition of knowledge distillation. Both works further use task identity information in some way to guide the generation, whereas we do not inform our model about task identity and use an unconditional diffusion model. In other recent work, \citet{smith2023continual} briefly report that they observe a rapid accumulation of errors when using standard generative replay, casting doubt on its efficacy in addressing catastrophic forgetting in diffusion models. To address this challenge, we propose a distillation-based approach that mitigates this issue and makes generative replay viable.



\section{Methods}\label{sec:method}
In this section, we first revisit standard generative replay for training diffusion models and explain how we implement it in our continual learning experiments. Second, we introduce our proposed approach, \textit{generative distillation}, which combines generative replay with knowledge distillation to more effectively make use of the information contained in the previous task model.

\subsection{Standard generative replay}\label{methods_generative_replay}
In the continual learning setting we consider, a diffusion model $\epsilon_{\theta}$ is incrementally trained on a sequence of tasks $\mathcal{T}=\{\mathcal{T}_1, \mathcal{T}_2, ..., \mathcal{T}_B\}$. Before starting to train on a new task~$\mathcal{T}_i$, we first store a copy of the diffusion model (i.e., a copy of the model as it is after finishing training on the previous task). We call this stored model copy the `previous task model' or `teacher', and denote it by~$\Teacher$. The model that is continued to be trained on the new task~$\mathcal{T}_i$, we call the `current task model' or `student', and we denote it by~$\Student$. (Note that the current task model~$\Student$ is initialized from the previous task model~$\Teacher$, as is common practice in continual learning literature.)

With standard generative replay, when the current task model~$\Student$ is trained on task~$\mathcal{T}_i$, the previous task model~$\Teacher$ is used to generate replay samples which are combined with the training data of the current task. For every sample that is replayed, the previous task model~$\Teacher$ generates an image sample using DDIM, to which Gaussian noise~$\epsilon \sim \mathcal{N}(0, I)$ is added to obtain a noisy sample~$x_t$ at timestep~$t$:
\begin{equation}\label{eq:input_distillation_GR}
    x_t = \sqrt{\alpha_t} \cdot \text{DDIM}_N(\Teacher) + \sqrt{1-\alpha_t} \cdot \epsilon,
\end{equation} 
where $\text{DDIM}_N(\Teacher)$ is a clean sample generated by $\Teacher$ using DDIM with $N$~generation steps, and $\alpha$ is the noise schedule function. The loss on the replayed data is then:
\begin{equation}\label{eq:final_objective_gr}
    L_\text{replay}(\theta_i) = \Exp{}_{t \sim \text{Unif}(T), x_t \sim \Teacher} \bigl[ 
    || \epsilon - \Student(x_t, t)
    ||^2_2 \bigr].
\end{equation}
where $\epsilon$ is the noise added to the replayed image in \Cref{eq:input_distillation_GR}.

The final loss for generative replay is a weighted combination of the standard loss (\Cref{eq:gr_loss_generator_simple}) for data from the current task and the replay loss for replayed data: 
\begin{equation}\label{eq:gr_loss_generator}
    L_{\text{GR}} = r L_{\text{simple}} + (1-r) L_{\text{replay}},
\end{equation}
where $r$ is a ratio that gives the loss on the current task a decreasing proportional importance relative to the number of tasks seen so far (i.e., $r=\frac{1}{i}$ with $i$ the number of tasks learned so far) \citep{vandeven2020brain}.

\subsection{Generative distillation}
\label{sec:methods-generative-distillation}

\begin{figure}[t]
    \centering
    \begin{subfigure}[b]{0.47\textwidth}
         \centering
         \includegraphics[width=\textwidth]{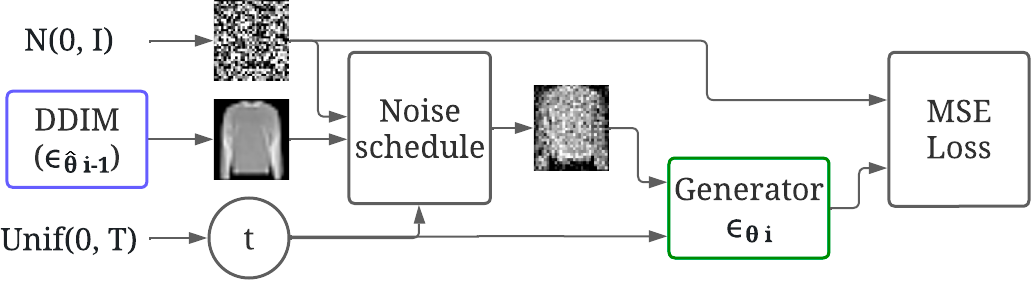}
         \caption{Generative Replay}
         \label{fig:gen_replay}
    \end{subfigure}
    \qquad
    \begin{subfigure}[b]{0.47\textwidth}
         \centering
         \includegraphics[width=\textwidth]{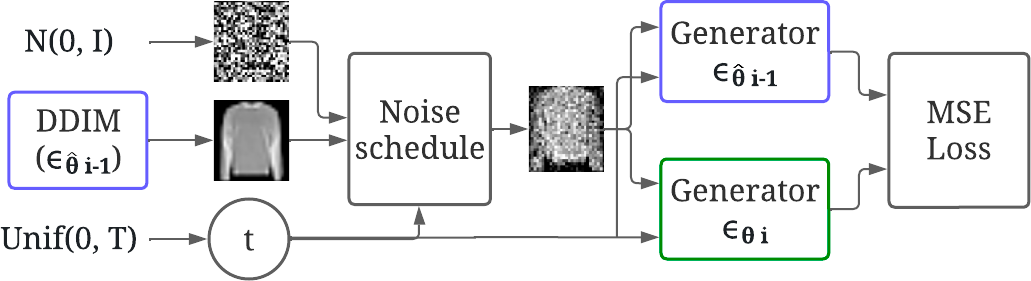}
         \caption{Generative Distillation}
         \label{fig:gen_distil}
    \end{subfigure}
    \caption{\textbf{Conceptual flow of generative replay and generative distillation.} In generative replay, the current task model $\Student$ is trained to match the noise added to the replay sample, whereas, in generative distillation, $\Student$ is trained to match the noise prediction of the previous task model $\Teacher$.}
    \label{fig:gr_distil}
\end{figure}

In standard generative replay, the previous task model $\Teacher$ is used in the training process of the current task model $\Student$ only for sampling images for replay (\Cref{fig:gen_replay}, see \Cref{methods_generative_replay} for details). This means that during training, knowledge from $\Teacher$ is transferred to $\Student$ only at the endpoint of the reverse process of the diffusion model, i.e.~after going through all the denoising steps. Therefore, if the previous task model generates imperfect images (e.g., using a relatively small number of DDIM steps), the current task model might learn trajectories that lead to these images, causing an accumulation of errors. To tackle this issue, we introduce knowledge distillation into generative replay for diffusion models. Our approach, \emph{generative distillation}, transfers the knowledge from~$\Teacher$ to~$\Student$ at each point of the reverse trajectory, while using synthetic images sampled from~$\Teacher$ as inputs. 

With generative distillation, similar as with generative replay, the previous task model~$\Teacher$ generates first an image sample using DDIM and, then, Gaussian noise is added to obtain a noisy sample~$x_t$ at timestep~$t$ as in \Cref{eq:input_distillation_GR}.
Given the noisy sample~$x_t$, the distillation objective for incorporating knowledge from previous tasks during generative distillation is the following:
\begin{equation}\label{eq:final_objective}
    L_\text{distil}(\theta_i) = \Exp{}_{t \sim \text{Unif}(T), x_t \sim \Teacher} \bigl[ ||\Teacher(x_t, t)-\Student(x_t, t)||^2_2 \bigr].
\end{equation}

In \Cref{fig:gen_distil}, a conceptual flow of generative distillation is shown. The final loss that we aim to minimize throughout the continual learning process is a weighted combination of the standard loss (\Cref{eq:gr_loss_generator_simple}) for data from the current task and the distillation loss for replayed data:
\begin{equation}\label{eq:gd_loss_generator}
    L_{\text{GD}} = r L_{\text{simple}} + (1-r) \lambda L_{\text{distil}},
\end{equation}
where $r$ is the same task importance ratio as in \Cref{eq:gr_loss_generator}, and $\lambda$ is a weighting parameter to balance the magnitudes of the standard and distillation losses.


\begin{algorithm}[t]
    \caption{Generative distillation}\label{alg:generative_distillation}
    \textbf{Input:} Training data current task~$\mathcal{D}_i$, previous task model~$\Teacher$, current task model~$\Student$, number of generation steps~$N$, and weighting parameter~$\lambda$. 
    \begin{algorithmic}[1]
    \For{$x_0 \in \mathcal{D}_i$}
        \State Sample $t \sim \text{Unif}({1,...,T})$, $s \sim \text{Unif}({1,...,T})$
        \State Sample $\epsilon \sim \mathcal{N}(0, I)$, $\bar{\epsilon} \sim \mathcal{N}(0, I)$
        \State $x_t = \sqrt{\alpha_t} x_0 + \sqrt{1-\alpha_t} \epsilon$
        \State $\bar{x}_s = \sqrt{\alpha_s} \cdot \text{DDIM}_N(\Teacher) + \sqrt{1-\alpha_s} \bar{\epsilon}$
        \State Take gradient descent on 
        \State \hspace{\algorithmicindent} $\nabla_{\theta_i} \big[ \frac{1}{i} ||\epsilon - \Student(x_t, t)||^2_2 + (1-\frac{1}{i}) \lambda ||\Teacher(\bar{x}_s, s)-\Student(\bar{x}_s, s)||^2_2$ \big]
    \EndFor
    \end{algorithmic}
\end{algorithm}

\section{Experiments}\label{sec:experiments}

To compare the effectiveness of standard generative replay and our proposed generative distillation approach for continually training a diffusion model, we perform experiments using the Fashion-MNIST \citep{xiao2017fashion} and CIFAR-10 \citep{krizhevsky2009learning} datasets. We divide both datasets into five tasks, each comprising two randomly selected classes (\Cref{fig:task_order_cifar}), that must be learned one after the other.
For both task sequences, the diffusion model that we train is based on the U-Net architecture used for CIFAR-10 in \citet{ho2020denoising} (refer to \Cref{appendix:models_diffusion} for details).
When using generative replay or generative distillation, we use the previous task model~$\Teacher$ to generate new samples to replay at each training step. We always generate as many replay samples as there are samples from the current task. To generate these replay samples, we use 2~DDIM steps on Fashion-MNIST and 10~DDIM steps on CIFAR-10. We base this choice on samples generated with a model trained on an independently and identically distributed stream of data using different amounts of DDIM steps (refer to \Cref{appendix:samples_joint_data}). For each dataset, we select the smallest number of DDIM steps that provide distinguishable objects. For further details on the experiments, models and metrics, please refer to \Cref{appendix:extra_details}.

In the following, we mainly run experiments on the strategies explained in the previous section and baselines:
\setlist{nolistsep}
\begin{itemize}[noitemsep]
    \item \textbf{Generative replay}: the previous task model generates replay samples for the current task model. 
    \item \textbf{Generative distillation}: the previous task model generates replay samples \emph{and} transfers knowledge through distillation to the current task model. 
    \item \textbf{Joint}: a model trained on data from all tasks, serving as our upper target. 
    \item \textbf{Finetune}: a model naïvely fine-tuned on the current task data, serving as our lower target.
\end{itemize}

\subsection{Generative replay results in a catastrophic loss in denoising capability}
\label{sec:catastrophic}
\begin{figure}[t]
\centering
\begin{subfigure}{\textwidth}
\begin{subfigure}[b]{0.195\textwidth}
     \centering
     \includegraphics[width=\textwidth]{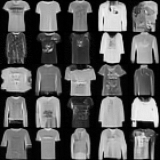}
\end{subfigure}
\begin{subfigure}[b]{0.195\textwidth}
     \centering
     \includegraphics[width=\textwidth]{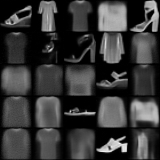}
\end{subfigure}
\begin{subfigure}[b]{0.195\textwidth}
     \centering
     \includegraphics[width=\textwidth]{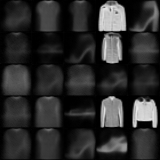}
\end{subfigure}
\begin{subfigure}[b]{0.195\textwidth}
     \centering
     \includegraphics[width=\textwidth]{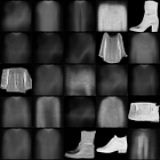}
\end{subfigure}
\begin{subfigure}[b]{0.195\textwidth}
     \centering
     \includegraphics[width=\textwidth]{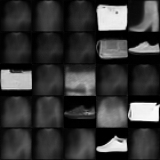}
\end{subfigure}
\vspace{8pt}
\setcounter{subfigure}{0}
\end{subfigure}
\begin{subfigure}{\textwidth}
\begin{subfigure}[b]{0.195\textwidth}
     \centering
     \includegraphics[width=\textwidth]{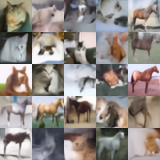}
     \caption{Task 1}
\end{subfigure}
\begin{subfigure}[b]{0.195\textwidth}
     \centering
     \includegraphics[width=\textwidth]{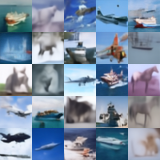}
     \caption{Task 2}
\end{subfigure}
\begin{subfigure}[b]{0.195\textwidth}
     \centering
     \includegraphics[width=\textwidth]{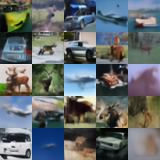}
     \caption{Task 3}
\end{subfigure}
\begin{subfigure}[b]{0.195\textwidth}
     \centering
     \includegraphics[width=\textwidth]{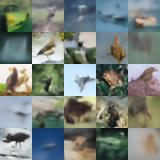}
     \caption{Task 4}
\end{subfigure}
\begin{subfigure}[b]{0.195\textwidth}
     \centering
     \includegraphics[width=\textwidth]{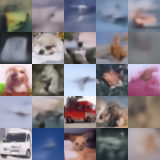}
     \caption{Task 5}\label{fig:experiments_qualitative_gr_last}
\end{subfigure}
\end{subfigure}
\caption{\textbf{Samples for generative replay on Fashion-MNIST (first row) and CIFAR-10 (second row).} To generate samples for previous tasks, the teacher used 2 DDIM steps on Fashion-MNIST and 10 DDIM steps on CIFAR-10. Standard generative replay causes a catastrophic accumulation of error for samples from previous tasks.}
\label{fig:experiments_qualitative_gr}
\end{figure}

The images generated by a diffusion model are imperfect approximations of the true data distribution that present some error, even if they look decent and are distinguishable to the human eye. As shown in \Cref{fig:experiments_qualitative_gr}, generative replay causes the error to add up, leading to a steep and steady degradation of the generation quality. The images generated by the diffusion model after training with generative replay become smooth and poorly detailed. 
According to \citet{deja2022analyzing}, diffusion models can be decomposed in two parts: (i) a generative part that generates an image from corrupted noise, and (ii) a denoising part that seems responsible for also introducing new details to the image. In the continuous learning process, the model seems to maintain its generative capabilities, at least to some extent. For instance, when examining the second column in the last row of Fashion-MNIST in \Cref{fig:experiments_qualitative_gr}, the model appears to preserve the t-shirt silhouette despite an increase in noise and a severe loss of finer details. Similarly, in CIFAR-10, the last sample in the final row exhibits a comparable behavior, where, in the last task (refer to \Cref{fig:experiments_qualitative_gr_last}), only an extremely blurry version of the original cat sample remains. These observations suggest that in particular the denoising capabilities of a diffusion model are susceptible to the shortcomings of generative replay. 

\subsection{Generative distillation substantially improves upon generative replay}
\label{sec:exps-generative-distillation}

In the previous subsection, we observed that standard generative replay with a limited number of DDIM steps causes a severe degradation of the diffusion model's denoising capabilities. Now we explore whether, without increasing the number of DDIM steps, this challenge can be overcome by our proposed generative distillation approach.

\subsubsection{Qualitative comparison}

\begin{figure}
\centering
\begin{subfigure}{\textwidth}
\begin{subfigure}[b]{0.25\textwidth}
     \centering
     \includegraphics[width=\linewidth]{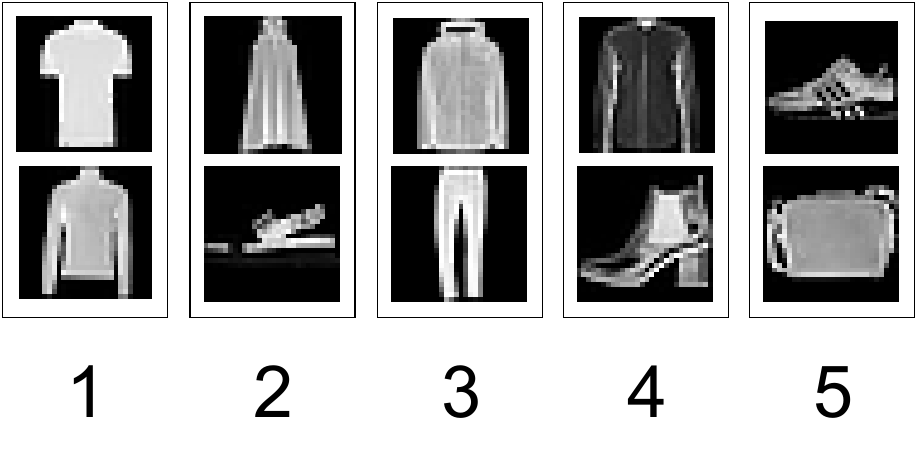}
\end{subfigure}
\hspace{0.05\textwidth}
\begin{subfigure}[b]{0.223\textwidth}
     \centering
     \includegraphics[width=\textwidth]{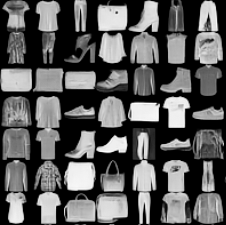}
\end{subfigure}
\begin{subfigure}[b]{0.223\textwidth}
     \centering
     \includegraphics[width=\textwidth]{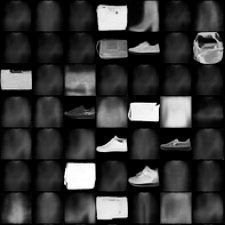}
\end{subfigure}
\begin{subfigure}[b]{0.223\textwidth}
     \centering
     \includegraphics[width=\textwidth]{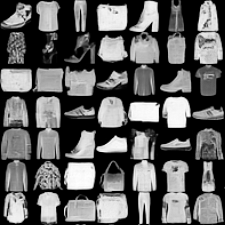}
\end{subfigure}
\vspace{0.15in}
\setcounter{subfigure}{0}
\end{subfigure}
\begin{subfigure}{\textwidth}
\begin{subfigure}[b]{0.25\textwidth}
     \centering
     \includegraphics[width=\linewidth]{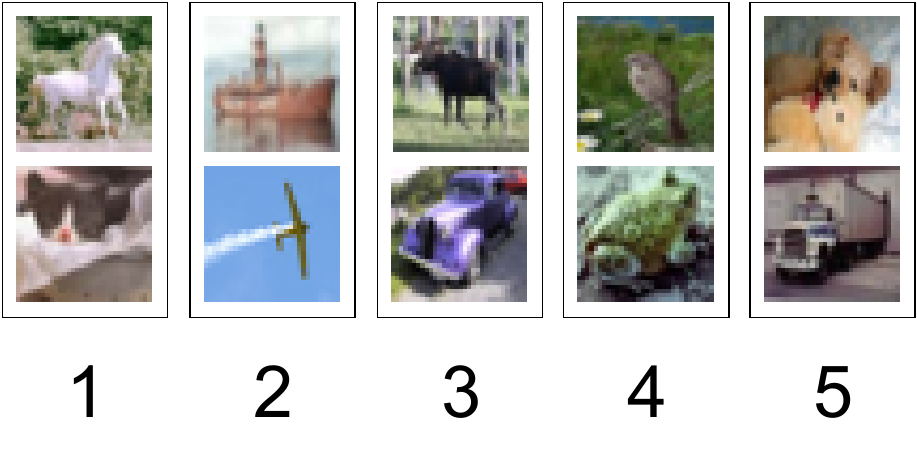}
     \caption{Task order}
     \label{fig:task_order_cifar}
\end{subfigure}
\hspace{0.05\textwidth}
\begin{subfigure}[b]{0.223\textwidth}
     \centering
     \includegraphics[width=\textwidth]{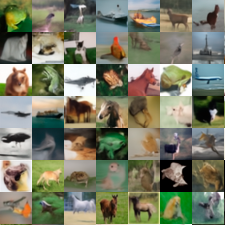}
     \caption{Joint}
\end{subfigure}
\begin{subfigure}[b]{0.223\textwidth}
     \centering
     \includegraphics[width=\textwidth]{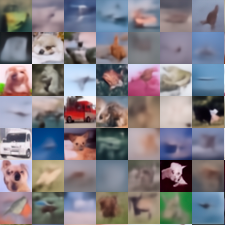}
     \caption{GR}
     \label{fig:results_qualitative_gr_cifar}
\end{subfigure}
\begin{subfigure}[b]{0.223\textwidth}
     \centering
     \includegraphics[width=\textwidth]{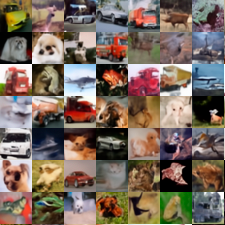}
     \caption{GD}
     \label{fig:results_qualitative_gd_cifar}
\end{subfigure}
\end{subfigure}
\caption{\textbf{Qualitative comparison of generative replay and generative distillation.} The first row corresponds to results on Fashion-MNIST and the second row on CIFAR-10. (a)~The order of the tasks the models were trained on. (b-d)~Images generated at the end of training by a diffusion model jointly trained on all tasks (Joint), continually trained using generative replay (GR) or continually trained using generative distillation (GD). Introducing distillation into generative replay prevents the catastrophic accumulation of error and the model is still able to produce detailed images.}
\label{fig:results_qualitative}
\end{figure}

In \Cref{fig:results_qualitative}, we show images sampled from the diffusion model after training on all tasks. 
While the model trained with generative replay generated smooth and blurry images, the model trained with generative distillation was still able to produce detailed images on both datasets. On top of that, the samples from generative distillation are visually comparable to those obtained after joint training, that is, after sequentially training the diffusion model using all the data from tasks so far. This demonstrates the effectiveness of generative distillation in preserving past knowledge.

\subsubsection{Quantitative comparison}\label{sec:exp_quantitative_comparison}
To quantitatively compare the trained diffusion models, we report the commonly used Fréchet Inception Distance~(FID)~\citep{heusel2017gans}. We also report the Kullback-Leibler divergence (KLD) between the ground truth class frequency distribution and the distribution of predicted class membership for images generated by the diffusion model. The KLD evaluates to what extent the diffusion model generates samples from all classes seen so far (rather than, for example, only from classes of the last task). 
When computing these evaluation metrics, we sample images using 10~DDIM steps on Fashion-MNIST and 20 DDIM steps on CIFAR-10. See \Cref{apx:evaluation} for details.

\begin{table}[t]
    \centering
    \caption{\label{tab:results_cl_gen}\textbf{Quantitative comparison of generative replay and generative distillation.} These quantitative results confirm that introducing distillation into generative replay substantially improves the continual learning performance of diffusion models. The FID and KLD are computed after training on all tasks. Reported for each metric is mean ± standard error over 3 random seeds.}
    \begin{tabular}{llllll}
\toprule
& \multicolumn{2}{c}{\bf Fashion MNIST} & & \multicolumn{2}{c}{\bf CIFAR-10} \\
Strategy & ~~~~$\downarrow \text{FID}$ & ~~$\downarrow \text{KLD}$ & & ~~~$\downarrow \text{FID}$ & ~~$\downarrow \text{KLD}$ \\
\midrule
    Joint \scriptsize{(upper target)} & ~~15.5 \scriptsize{$\pm$ 1.3} & 0.08 \scriptsize{$\pm$ 0.02} & & 31.9 \scriptsize{$\pm$ 2.9} & 0.12 \scriptsize{$\pm$ 0.02}\\
    Finetune \scriptsize{(lower target)} & ~~68.8 \scriptsize{$\pm$ 7.0} & 4.82 \scriptsize{$\pm$ 1.42} & & 55.8 \scriptsize{$\pm$ 2.6} & 1.14 \scriptsize{$\pm$ 0.10}\\
    \midrule
    Generative replay & 211.8 \scriptsize{$\pm$ 6.9} & 3.39 \scriptsize{$\pm$ 0.30} & & 96.4 \scriptsize{$\pm$ 5.3} & 0.74 \scriptsize{$\pm$ 0.14}\\
    Generative distillation & ~~\textbf{23.5 \scriptsize{$\pm$ 1.1}} & \textbf{0.25 \scriptsize{$\pm$ 0.10}} & & \textbf{42.3 \scriptsize{$\pm$ 4.7}} & \textbf{0.15 \scriptsize{$\pm$ 0.01}} \\
    \bottomrule  
\end{tabular}
\end{table}

\begin{figure}
    \vspace{-2pt}
    \centering

    \begin{subfigure}{\textwidth}
    \begin{subfigure}[b]{0.471\textwidth}
        \centering
        \includegraphics{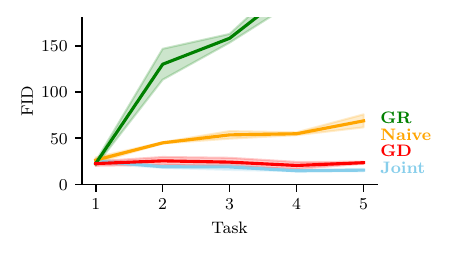}
    \end{subfigure}
    \hfill
    \begin{subfigure}[b]{0.471\textwidth}
        \centering
        \includegraphics{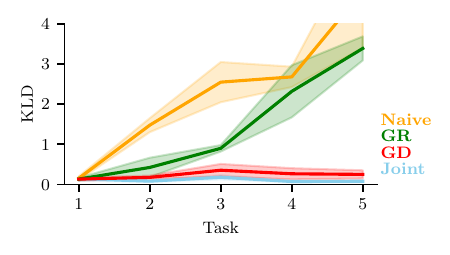}
    \end{subfigure}
    \vspace{-0.1in}
    \caption{Fashion-MNIST}
    \end{subfigure}
    
    \begin{subfigure}{\textwidth}
    \begin{subfigure}[b]{0.471\textwidth}
        \centering
        \includegraphics{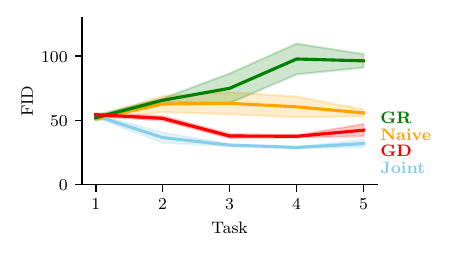}
    \end{subfigure}
    \hfill
    \begin{subfigure}[b]{0.471\textwidth}
        \centering
        \includegraphics{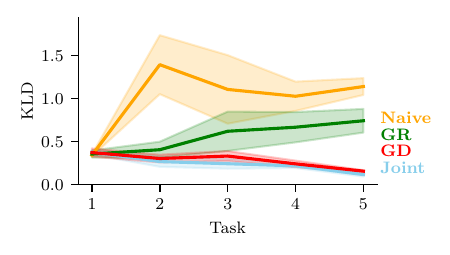}
    \end{subfigure}
    \vspace{-0.1in}
    \caption{CIFAR-10}
    \end{subfigure}
    
    \caption{\textbf{Quality metrics of the diffusion model throughout the continual learning process.} For each approach, the FID and KLD of the diffusion model are computed after every task. Displayed are the means ± standard errors over 3 random seeds. Joint: training using data from all tasks so far (upper target), Naive: continual fine-tuning (lower target), GR: generative replay, GD: generative distillation.}
    \label{fig:results_cl_metrics_through_tasks_combined}
\end{figure}

The results in \Cref{tab:results_cl_gen} and \Cref{fig:results_cl_metrics_through_tasks_combined} back up the observations made in the qualitative assessment. For example, the use of generative distillation resulted in $9.01\times$ lower FID on Fashion MNIST and $2.28\times$ lower FID on CIFAR-10 compared to generative replay. Similar results were observed for the KLD, which is $13.56\times$ lower on Fashion MNIST and $4.93\times$ lower on CIFAR-10. 
These results show that generative distillation preserves the diversity of the diffusion model, achieving a KLD close to joint training. 
In the Appendix, we further show that generative distillation clearly outperforms generative replay also when increasing or decreasing the number of DDIM teacher steps (\Cref{fig:qualitative_gr2,fig:qualitative_gd2,fig:qualitative_gr10,fig:qualitative_gd10,fig:qualitative_gr100,fig:qualitative_gd100,fig:qualitative_gr10_cifar,fig:qualitative_gd10_cifar,fig:qualitative_gr100_cifar,fig:qualitative_gd100_cifar}).
In \Cref{appendix:computational_cost} we further compare the computational and memory costs of generative distillation and generative replay, and in \Cref{appendix:experience_replay} we compare generative distillation with experience replay using various buffer sizes.


\subsubsection{Evaluation with a continually trained classifier}
\label{sec:classifier}

\begin{table}[t]
    \centering
    \caption{\textbf{Evaluation with a continually trained classifier.} The classifier is trained on the data of the current task along with replay samples from the continually trained diffusion model. Reported is mean ± standard error over 3~seeds.}\label{tab:results_cl_clf}
    \begin{tabular}{lclc}
        \toprule
        & \textbf{Fashion-MNIST} & & \textbf{CIFAR-10} \\
        Strategy & $\uparrow \text{ACC}$ & & $\uparrow \text{ACC}$ \\
        \midrule
        Joint \it \scriptsize{(upper target)} & 0.92\scriptsize{ $\pm$ 0.00} & & 0.77\scriptsize{ $\pm$ 0.03} \\
        Finetune \it \scriptsize{(lower target)} & 0.17\scriptsize{ $\pm$ 0.03} & & 0.19\scriptsize{ $\pm$ 0.00} \\
        \midrule
        Generative replay & 0.50\scriptsize{ $\pm$ 0.04} & & 0.26\scriptsize{ $\pm$ 0.01} \\
        Generative distillation & \textbf{0.72\scriptsize{ $\pm$ 0.03}} & & \textbf{0.31\scriptsize{ $\pm$ 0.03}} \\
        \bottomrule
    \end{tabular}
\end{table}

As an additional way of evaluating the performance of the continually trained diffusion model, we include a classifier trained with generative replay using the continually trained diffusion model as the generator (see \Cref{appendix:cont_class} for details). When generating samples to train the classifier, we used 10 DDIM steps on Fashion-MNIST and 20 DDIM steps on CIFAR-10. In \Cref{tab:results_cl_clf}, we report the accuracy of the classifier at the end of training. In line with the results discussed in previous sections, the model trained with generative distillations proved to generate samples that are more useful for training the classifier in both datasets. The low accuracy achieved on CIFAR-10 may be explained by the higher complexity of the dataset and the relatively small amount of generation steps used to produce the samples. Nevertheless, this evaluation confirms the benefit of incorporating distillation into generative replay.

\subsection{Influence of the quality of the replayed inputs on generative distillation}\label{experiments:extra}

\begin{figure}
\centering

\begin{subfigure}{\textwidth}
\centering
\begin{subfigure}[b]{1pt}
\begin{adjustbox}{rotate=90, right}
\lapbox[0pt]{1.125\width}{Generative replay}
\end{adjustbox}
\end{subfigure}
\hspace{4pt}
\begin{subfigure}[b]{0.195\textwidth}
     \centering
     \includegraphics[width=\textwidth]{assets/experiment/gr2/0004_cropped.png}
\end{subfigure}
\begin{subfigure}[b]{0.195\textwidth}
     \centering
     \includegraphics[width=\textwidth]{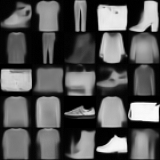}
\end{subfigure}
\begin{subfigure}[b]{0.195\textwidth}
     \centering
     \includegraphics[width=\textwidth]{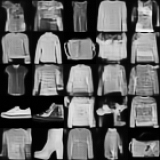}
\end{subfigure}

\vspace{8pt}

\begin{subfigure}{\textwidth}
\centering
\begin{subfigure}[b]{1pt}
\begin{adjustbox}{rotate=90, right}
\lapbox[0pt]{1.185\width}{Generative distillation}
\end{adjustbox}
\end{subfigure}
\hspace{4pt}
\begin{subfigure}[b]{0.195\textwidth}
     \centering
     \includegraphics[width=\textwidth]{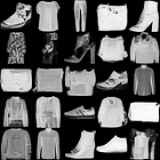}
     \caption{2 teacher steps}
\end{subfigure}
\begin{subfigure}[b]{0.195\textwidth}
     \centering
     \includegraphics[width=\textwidth]{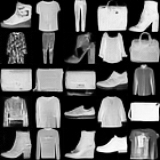}
     \caption{10 teacher steps}
\end{subfigure}
\begin{subfigure}[b]{0.195\textwidth}
     \centering
     \includegraphics[width=\textwidth]{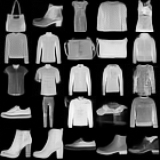}
     \caption{100 teacher steps}
\end{subfigure}
\vspace{0.1in}
\setcounter{subfigure}{0}
\end{subfigure}

\end{subfigure}
\caption{\textbf{Samples for different amounts of teacher steps on Fashion-MNIST.} Generative distillation continues to outperform generative replay when the number of DDIM steps to generate replay samples for previous tasks is varied. For a quantitative evaluation, see \Cref{tab:results_cl_full} in the Appendix.}
\label{fig:experiments_different_inputs}
\end{figure}

We studied the effect of increasing the number of DDIM steps used by the previous task model (teacher steps) in generative replay and generative distillation. As shown in \Cref{fig:experiments_different_inputs}, increasing the number of teacher steps in generative replay alleviated the catastrophic loss in denoising capabilities to some extent. However, generative replay continued to be consistently outperformed by generative distillation, even when using 100~teacher steps. Using 100 teacher steps in generative distillation did not seem to provide significant benefits over using 10 teacher steps, despite the greater computational costs. 

\begin{figure}
\centering
\begin{subfigure}{\textwidth}
\centering
\begin{subfigure}[b]{0.195\textwidth}
     \centering
     \includegraphics[width=\textwidth]{assets/experiment/gr2/0004_cropped.png}
\end{subfigure}
\begin{subfigure}[b]{0.195\textwidth}
     \centering
     \includegraphics[width=\textwidth]{assets/experiment/gd2/0004_cropped.png}
\end{subfigure}
\begin{subfigure}[b]{0.195\textwidth}
     \centering
     \includegraphics[width=\textwidth]{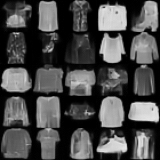}
\end{subfigure}
\begin{subfigure}[b]{0.195\textwidth}
     \centering
     \includegraphics[width=\textwidth]{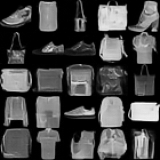}
\end{subfigure}

\vspace{8pt}

\setcounter{subfigure}{0}
\end{subfigure}
\begin{subfigure}{\textwidth}
\centering
\begin{subfigure}[b]{0.195\textwidth}
     \centering
     \includegraphics[width=\textwidth]{assets/experiment/cifar10/gr10/0004_cropped.png}
     \caption{Generative replay}
\end{subfigure}
\begin{subfigure}[b]{0.195\textwidth}
     \centering
     \includegraphics[width=\textwidth]{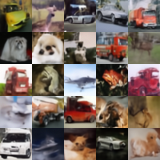}
     \caption{Generative distillation}
\end{subfigure}
\begin{subfigure}[b]{0.195\textwidth}
     \centering
     \includegraphics[width=\textwidth]{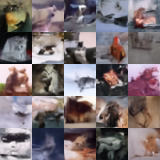}
     \caption{LwF}\label{fig:experiments_different_inputs_lwf}
\end{subfigure}
\begin{subfigure}[b]{0.195\textwidth}
     \centering
     \includegraphics[width=\textwidth]{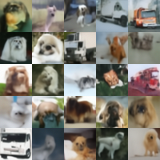}
     \caption{Gaussian}\label{fig:experiments_different_inputs_gaussian}
\end{subfigure}

\end{subfigure}
\caption{\textbf{Samples for different distillation strategies on Fashion-MNIST (first row) and CIFAR-10 (second row).} To generate samples for previous tasks, the teacher used 2 DDIM steps on Fashion-MNIST and 10 DDIM steps on CIFAR-10. Distilling with current data or Gaussian noise mitigates the catastrophic loss in denoising capabilities.}
\label{fig:experiments_different_inputs_strategies}
\end{figure}

We also compare generative distillation with other forms of distillation. One alternative we implemented distils the knowledge of the model from the previous task using the current task's data. This can be seen as a Learning without Forgetting~(LwF)~\citep{li2017learning} like approach applied to diffusion models. In \Cref{fig:experiments_different_inputs_lwf}, we can see that this form of distillation exhibits better performance than standard generative replay and mitigates the extent of the catastrophic loss in denoising capabilities. However, the overall quality of the samples degrades steadily and more artefacts show up as more tasks are learned (e.g.,~the sample in the fourth column of the last row of Fashion-MNIST or the first sample in the last row of CIFAR-10).

Distilling with Gaussian noise as input for all timesteps is another distillation approach we considered, which is an extreme case equivalent to the model from the previous task using 0 DDIM steps. Strikingly, this approach also outperforms generative replay as observed in \Cref{fig:experiments_different_inputs_gaussian}. Moreover, this strategy does not introduce as many artifacts as using current data (see the first sample in the last row of CIFAR-10). However, both approaches suffer a loss in diversity and tend to generate more samples from the current task, which could stem from a potential misalignment between the inputs and the true distribution of images.

\section{Limitations and future work}
One limitation of this work involves the way we searched for the lambda hyperparameter that balances the magnitudes of the standard loss and the distillation loss in \Cref{eq:gd_loss_generator}. The sweep that we performed for this hyperparameter is not part of the continual learning process, since we trained on the entire task sequence several times and selected the best one at the end (see \Cref{appendix_experimental_setup}). However, it is reassuring that the lambda values that we selected for Fashion MNIST also result in a strong performance for CIFAR-10.

Another factor to consider is that both generative replay and generative distillation introduce a significant computational overhead compared to finetuning, primarily due to the need for sample generation. Related to this, due to our limited time and computational resources, we used a number of training iterations that could be considered relatively small for diffusion models. For example, while in \citet{ho2020denoising} their models take 800k gradient updates on CIFAR-10, in our continual learning experiments, the models went through a total of 157k updates after training on all tasks. 

Restrictions on available resources also informed our choice to use relatively small-scale image datasets (Fashion MNIST and CIFAR-10) for the experiments. By using these datasets we were able to perform extensive and carefully controlled comparisons. Performing such controlled continual learning experiments with diffusion models on larger-scale benchmarks would not have been feasible for us. Given the consistent and substantial benefit that we found for generative distillation, combined with our conceptual arguments about the benefits of this approach, we expect that generative distillation will also offer benefits when used in larger-scale settings. This however remains to be demonstrated, which is something we leave for future work.

Although in this work we focused on using generative distillation with DDIM, we note that our approach can also be combined with other, newer samplers for diffusion models (e.g.,~\citealp{lu2022dpmsolver, lu2023dpmsolver, song2023consistency}). Such samplers might be able to allow generating samples for generative distillation with fewer teacher steps while still providing a good enough approximation of the true sample distribution. This would be particularly useful for more complex datasets, as we have observed that more teacher steps might be needed as the complexity of the dataset increases.

\section{Conclusions}
Continually training a diffusion model with standard generative replay results in a catastrophic loss in its denoising capabilities for past tasks, leading the model to produce blurry images. To address this problem, we propose including knowledge distillation into the generative replay process (generative distillation). Our experiments on Fashion-MNIST and CIFAR-10 show that generative distillation mitigates this catastrophic forgetting and markedly enhances the performance of the continually trained diffusion model. These results demonstrate the feasibility of continual learning in diffusion models, paving the way for future research into more robust, adaptable, and efficient generative models.

\section*{Acknowledgements}
This project has been supported by funding from the European Union under the Erasmus+ Traineeship program, under the Horizon 2020 research and innovation program (ERC project KeepOnLearning, grant agreement No.~101021347) and under Horizon Europe (Marie Skłodowska-Curie fellowship, grant agreement No.~101067759).

\bibliographystyle{collas2024_conference}
\bibliography{main}

\clearpage
\appendix
\renewcommand\thefigure{\thesection.\arabic{figure}}  
\renewcommand\thetable{\thesection.\arabic{table}}

\section{Extra details on the experiments}\label{appendix:extra_details}
\setcounter{figure}{0}
\setcounter{table}{0}

Code is publicly available at \href{https://github.com/Atenrev/diffusion_continual_learning}{https://github.com/Atenrev/diffusion\_continual\_learning}.

\subsection{Model architectures}

\subsubsection{Diffusion model}\label{appendix:models_diffusion}
For the experiments on Fashion-MNIST, we employ a custom 2D U-Net based on the one used in \citet{ho2020denoising}. We set the input size to $32 \times 32$ pixels, with a single input channel and a single output channel. We use four decreasing feature map resolutions ($32 \times 32$ to $4 \times 4$) with progressively increasing output channel dimensions: 64, 128, 256, and 512, respectively, each block containing one layer. The normalization is applied with 32 groups, and the model has two convolutional residual blocks per resolution level and self-attention blocks between the convolutional blocks at resolutions $8 \times 8$ and $4 \times 4$. The total amount of parameters is 42.9 million. 

For the experiments in CIFAR-10, we use the same architecture used in \citet{ho2020denoising} for the same dataset without any customization.

\subsubsection{Continually trained classifier}
\begin{table}[htbp]
    \caption{Architecture used for the continually trained classifier.}
    \label{tab:app_classifier_arch}
    \centering
    \begin{tabular}{|c|c|c|c|c|}
        \hline
        Layer & Type & Filters & Kernel Size & Activation \\
        \hline
        1 & Conv2D & 32 & 3x3 & ReLU \\
        2 & Conv2D & 32 & 3x3 & ReLU \\
        3 & MaxPool2D & - & 2x2 & - \\
        4 & Dropout (p=0.25) & - & - & - \\
        5 & Conv2D & 64 & 3x3 & ReLU \\
        6 & Conv2D & 128 & 3x3 & ReLU \\
        7 & MaxPool2D & - & 2x2 & - \\
        8 & Dropout (p=0.25) & - & - & - \\
        9 & Conv2D & 128 & 1x1 & ReLU \\
        10 & AdaptiveMaxPool2D & 1 & - & - \\
        11 & Dropout (p=0.25) & - & - & - \\
        12 & Linear & 10 & - & - \\
        \hline
    \end{tabular}
\end{table}

\subsubsection{Classifier used for computing the KLD metric}\label{appendix:models_kldclassifier}
In the implementation of the $\text{KLD}$ metric we use a ResNet18, which we had trained to classify the ten classes of Fashion-MNIST and CIFAR-10. For the training of this model on Fashion-MNIST, we split the training set into a training set consisting of 50,000 samples and a validation set consisting of 10,000. On CIFAR-10, we split the training set into a training set consisting of 45,000 samples and a validation set consisting of 5,000. We evaluated the model's accuracy on the validation set at the end of each epoch. Upon completing the training process, we chose the model from the epoch where it achieved the best validation accuracy. We applied the following augmentations, hyperparameters, and architectural changes:

\noindent\textbf{Augmentations:}
\setlist{nolistsep}
\begin{itemize}
\itemsep0em 
    \item Randomly resized the images to 32x32 pixels.
    \item Applied random horizontal flips to the images.
    \item Applied random rotations up to 10 degrees.
    \item Applied random affine transformations with a translation of up to 0.1 in both directions.
    \item Added Gaussian noise with a mean of 0 and a standard deviation of 0.1 to the images.
\end{itemize}

\noindent\textbf{Hyperparameters:}
\begin{itemize}
    \item Batch size: 64
    \item Number of training epochs: 100
    \item Learning rate: 0.0005
\end{itemize}

\begin{samepage}
\noindent\textbf{Architectural change to ResNet18:}
\begin{itemize}
    \item Replaced the final fully connected layer with a new linear layer with the number of output classes set to 10.
\end{itemize}
\end{samepage}

\subsection{Generative replay in the classifier}
\label{appendix:cont_class}
We continually train a classifier with the architecture displayed in \Cref{tab:app_classifier_arch} along with the diffusion model. This provides an additional way to evaluate the performance of the diffusion model when used as a generator in generative replay of a classifier~\citep{shin2017}. If the generator was ideal and approximated the real data distribution accurately, generative replay would be close to performing experience replay with an unlimited replay buffer.

For tasks $T_i, i>1$, we train the classifier $S_i$ on samples from the current task $T_i$, interleaved with synthetic samples generated by $\Teacher$, a copy of the knowledge-accumulating generator after training on the previous task. To generate the replay samples to train the classifier, we always use 10 DDIM steps on Fashion-MNIST and 20 DDIM steps on CIFAR-10. Since we use an unconditional diffusion model, we employ soft labelling, following \citet{vandeven2020brain}, which has shown greater effectiveness than hard labelling. These soft labels result from running the synthetic samples through the previous classifier $S_{i-1}$. Our objective is to minimize two distinct losses: one for the current task and one for previous tasks. The first loss $L_{\text{curr}}^{\text{clf}}$ is the usual cross-entropy classification defined as:
\begin{equation}\label{eq:gr_loss_classifier_curr}
    L_{\text{curr}}^{\text{clf}} = \mathbb{E}_{(x_{\text{curr}},y) \sim D_i} \biggr[  - \sum_{c=1}^{N_{\text{classes}}} y_c \log p_\theta(Y=c|x_{\text{curr}}) \biggl].
\end{equation}
The second loss is the distillation loss that makes use of soft labels defined as:
\begin{equation}\label{eq:gr_loss_classifier_replay}
    L_{\text{\text{replay}}}^{\text{clf}} = \mathbb{E}_{x_{\text{prev}} \sim \epsilon_{\theta_{i-1}}, \Tilde{y} \sim S_{i-1}} \biggr[ -R^2 \sum_{c=1}^{N_{\text{classes}}}{\Tilde{y}_c} \log p_\theta^R(Y=c|x_{\text{prev}}) \biggl]
\end{equation}
where $x_{\text{prev}}$ is the input data for the current task labelled with a hard target $y$, $x_{\text{prev}}$ is the input data for previous tasks generated by the previous generator, $\Tilde{y}$ are the soft targets for $x_{\text{prev}}$, and $R$ is a temperature parameter that we set to $2$.

The final objective is the joint loss that combines both with a ratio of relative importance $r=\frac{1}{i}$, with $i$ the number of tasks so far, that gives $L_{\text{replay}}$ an importance proportional to the number of tasks seen so far:
\begin{equation}\label{eq:gr_loss_classifier}
    L_{\text{classifier}} = r L_{\text{curr}}^{\text{clf}} + (1-r) L_{\text{replay}}^{\text{clf}}
\end{equation}

The experimental setup is discussed in \Cref{appendix_experimental_setup}.

\subsection{Additional distillation approaches}\label{appendix:additional_distillation}
The implementation of the distillation with current data is based on \Cref{eq:gd_loss_generator}, except that in $L_\text{distil}$ in \Cref{eq:final_objective}, $x_t$ is equal to the same noisy data from the current batch used in the standard training loss $x_t: \sqrt{\alpha_t} \cdot x_0 + \sqrt{1-\alpha_t} \cdot \epsilon$. Distillation with Gaussian noise is the same, except that now Gaussian noise is used as the ``noisy data input'', such that $x_t \sim \mathcal{N}(0, I)$ for all timesteps.

\subsection{Experimental setup}\label{appendix_experimental_setup}
In our experiment, we continually train a diffusion model along a classifier. When training on a new task, a copy of the diffusion model from previous tasks $\Teacher$ generates replay samples for the training of both the diffusion model (only on generative methods) and the classifier. To generate the replay samples to train the diffusion model, for all experiments reported in the main text, we use 2 DDIM steps on Fashion-MNIST and 10 DDIM steps on CIFAR-10. In the \Cref{appendix:extra_experiments}, we additionally report results for experiments in which we use 2, 10 or 100 DDIM steps to generate the replay samples to train the diffusion model. The replay samples to train the classifier are generated using 10 DDIM steps on Fashion-MNIST of 20 DDIM steps on CIFAR-10, see \Cref{appendix:cont_class}. No augmentations are applied to any of the samples. We standardize the ground truth data to fall within the interval of $[-0.5, 0.5]$ and resize it to $32x32$. 

For each task, the diffusion model is trained for 100 epochs (9,400 gradient updates) on Fashion-MNIST and 200 epochs (31,400 gradient updates) on CIFAR-10. The classifier is trained for 20 epochs (1,800 gradient updates). On Fashion-MNIST, the batch size is set to 128 for the first task and 256 for the subsequent ones to accommodate the synthetic samples (each batch then contains 128 samples from the current task and 128 replayed samples). On CIFAR-10, the batch size is set to 64 for the first task and 128 for the subsequent ones. For distillation with current data the batch size is always 128 and 64 for Fashion-MNIST and CIFAR-10 respectively. We train both models using Adam~\citep{kingma2017adam} as the optimizer. For the diffusion model, we use a learning rate of $2 \times 10^{-4}$, while for the classifier we use $1 \times 10^{-3}$. We implement all the strategies using the Avalanche library~\citep{lomonaco2021avalanche}.

In order to find the best $\lambda$ for each distillation approach, we sweep over different sets of lambda values based on the observed loss magnitudes. We sweep on Fashion-MNIST across 3 random seeds over the values of  $\{0.1, 0.25, 0.5, 0.75, 1.0, 1.25, 1.5, 1.75, 2.0, 2.25, 2.5, 2.75, 3.0\}$ for generative distillation,  $\{1, 4, 8, 12, 16, 20, 24, 28\}$ for distillation with Gaussian noise, and  $\{0.1, 0.25, 0.5, 0.75, 1.0, 1.25, 1.5, 1.75\}$ for distillation with current data. We select the lambda that yields the best FID value for the comparisons and use it for both Fashion-MNIST and CIFAR-10.

\subsection{Evaluation}
\label{apx:evaluation}
The models are evaluated after finishing the training on each task. For these evaluations, we only use the part of the test set corresponding to all the classes seen so far. To compute the Fréchet Inception distance~(FID), we generate 10,000 images with the diffusion model using 10 DDIM steps. We use these same 10,000 images to compute the Kullback-Leibler divergence (KLD) between the ground truth class frequency distribution and the generated class frequency distribution. Since we are using an unconditional diffusion model architecture, we use a ResNet18~\citep{he2015deep} trained on Fashion-MNIST or on CIFAR-10 to label the generated samples; see \Cref{appendix:models_kldclassifier} for further details on the training of this classifier. As an additional metric, we report the accuracy of the continually trained classifier with generative replay (see \Cref{appendix:cont_class} for details).

To visualize example images sampled from the trained diffusion models, we always use 10 DDIM steps on Fashion-MNIST and 20 DDIM steps on CIFAR-10.

\section{Extra experiments}\label{appendix:extra_experiments}
\setcounter{figure}{0}
\setcounter{table}{0}

\subsection{Samples from a model trained jointly on all data}\label{appendix:samples_joint_data}

To motivate the choice of the number of teacher DDIM steps, we trained a diffusion model in Fashion-MNIST and CIFAR-10 with an independently and identically distributed stream of data for 100 epochs (equivalent to 46,900 gradient updates) and 200 epochs (equivalent to 78,200 gradient updates) respectively. We set the batch size to 128 and the learning rate to $2 \times 10^{-4}$. With this model, we used DDIM to generate images using 2, 10 and 100 steps. For comparison, we also sampled images using DDPM.

\paragraph{Fashion-MNIST}

\begin{figure}[ht]
\centering
\begin{subfigure}[b]{0.225\textwidth}
     \centering
     \includegraphics[width=\textwidth]{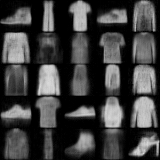}
\end{subfigure}
\begin{subfigure}[b]{0.225\textwidth}
     \centering
     \includegraphics[width=\textwidth]{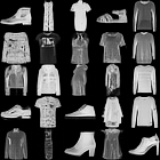}
\end{subfigure}
\begin{subfigure}[b]{0.225\textwidth}
     \centering
     \includegraphics[width=\textwidth]{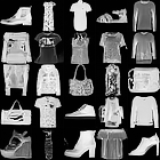}
\end{subfigure}
\begin{subfigure}[b]{0.225\textwidth}
     \centering
     \includegraphics[width=\textwidth]{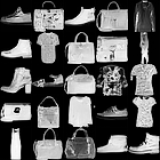}
\end{subfigure}

\begin{subfigure}[b]{0.225\textwidth}
     \centering
     \includegraphics[width=\textwidth]{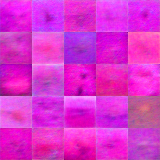}
     \caption{DDIM (2 steps)}
\end{subfigure}
\begin{subfigure}[b]{0.225\textwidth}
     \centering
     \includegraphics[width=\textwidth]{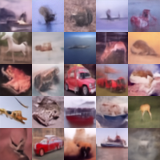}
     \caption{DDIM (10 steps)}
\end{subfigure}
\begin{subfigure}[b]{0.225\textwidth}
     \centering
     \includegraphics[width=\textwidth]{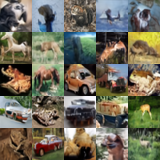}
     \caption{DDIM (100 steps)}
\end{subfigure}
\begin{subfigure}[b]{0.225\textwidth}
     \centering
     \includegraphics[width=\textwidth]{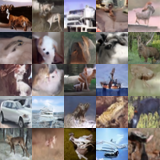}
     \caption{DDPM}
\end{subfigure}
\caption{\textbf{Samples from a jointly trained model on Fashion-MNIST (first row) and CIFAR-10 (second row) with different sampling methods.} Relatively smooth samples are generated with 2 DDIM steps, minor artefacts are present with 100 DDIM steps.}
\label{fig:samples_iid}
\end{figure}

As illustrated in \Cref{fig:samples_iid}, employing 2 DDIM steps resulted in a distinguishable but smooth approximation of the samples. Using 10 DDIM steps provided decent looking samples; increasing the DDIM step count to 100 did not seem to enhance the visual quality of the sampled images much.

\paragraph{CIFAR-10}

In contrast with the results obtained with the model trained on Fashion-MNIST, the model trained on CIFAR-10 produced noisy and undistinguishable samples when using 2 DDIM steps as shown in \Cref{fig:samples_iid}. When the number of DDIM steps is increased to 10, the objects become recognizable despite being a bit smooth. With 100 DDIM steps or when using DDPM, the images yielded by the model become sharper and more detailed.

\subsection{Computational and memory overhead}\label{appendix:computational_cost}

Compared with generative replay, generative distillation should increase the computational costs by just one forward pass. This extra forward pass is needed to compute $\Teacher(x_t, t)$ in \Cref{eq:final_objective}. To verify this in practice, here we compare the empirical wall time per iteration for finetuning, generative replay and generative distillation. To obtain the wall time per iteration for each method, we take the average over 1000 training steps on an NVIDIA GeForce 1080Ti GPU. As in our main experiments, when generating replay data in generative replay and generative distillation, we use 2~DDIM teacher steps with Fashion MNIST and 10~DDIM teacher steps with CIFAR-10. 
As shown in \Cref{tab:computational_results}, when comparing generative distillation with generative replay, the relative increase in wall time on Fashion MNIST (1.13×) and CIFAR-10 (1.06×) is moderate, especially considering the substantially lower FID achieved on Fashion MNIST (9.01×) and on CIFAR-10 (2.28×) (refer to \Cref{tab:results_cl_gen} in the main text).

We also report the memory requirements of each method. For finetuning, there is no need to store either samples or model copies; only the parameters of the model that is being trained need to be kept in memory (for Fashion MNIST: 42,855,873~parameters; for CIFAR-10: 35,746,307~parameters). For generative replay and generative distillation, there is no need to store a buffer of synthetic samples (since we generate the samples to be replayed at each iteration), but for both approaches one extra model copy needs to be stored (the previous task model~$\Teacher$). Therefore, generative replay and generative distillation have the same memory requirement, double that of finetuning (for Fashion MNIST: 85,711,746~parameters; for CIFAR-10: 71,492,614~parameters).

\begin{table}[]
    \centering
    \caption{\label{tab:computational_results}\textbf{Computational overhead for different approaches.} Reported is the empirical wall-time (in ms) per iteration (mean ± standard deviation over 1000 iterations).}
    \begin{tabular}{llll}
\toprule
& \multicolumn{1}{c}{\bf Fashion MNIST} & & \multicolumn{1}{c}{\bf CIFAR-10} \\
Strategy & $\downarrow$ wall-time (\text{ms/it}) & & $\downarrow$ wall-time (\text{ms/it}) \\
\midrule
    Finetune & ~~~~253.3 \scriptsize{$\pm$ 8.8} & & ~~~~~~383.4 \scriptsize{$\pm$ 95.3} \\
    Generative replay & ~~~~664.3 \scriptsize{$\pm$ 3.8} & & ~~~~1881.8 \scriptsize{$\pm$ 42.9} \\
    Generative distillation & ~~~~747.9 \scriptsize{$\pm$ 3.6} & &  ~~~~1992.7 \scriptsize{$\pm$ 47.6} \\
    \bottomrule  
\end{tabular}
\end{table}


\subsection{Comparing generative distillation with experience replay}\label{appendix:experience_replay}

We include additional experiments where we compare the performance of generative distillation with experience replay using various memory buffer sizes. We consider memory buffers with a total capacity of 10, 100, and 1000~samples. After each task, the capacity of the buffer is equally divided over all previous tasks (i.e.,~with a total capacity of ten samples, after the first task ten samples per task are stored, after the second task five samples per task are stored, and so on). The samples to be added to the memory buffer are randomly selected from the last task's training data, and the samples to be discarded from the buffer (to make place for the samples from the last task) are also selected at random. During training, in each iteration, the samples to be replayed are randomly sampled from the buffer. Similar to our implementation of generative replay, the size of the mini-batch of replay data matches that of the current task mini-batch (if the memory buffer size is smaller than the mini-batch size, all samples from the memory buffer are replayed). We compute a separate loss on the data of the current task and on the replayed data (using \Cref{eq:gr_loss_generator_simple} for both), and we combine these two losses as in \Cref{eq:gr_loss_generator}.

In \Cref{tab:results_cl_er}, the results indicate that increasing the buffer size generally improves the performance of experience replay. On Fashion MNIST, using a buffer with total capacity of 1000 achieves comparable performance to generative distillation in terms of FID and KLD. On CIFAR-10, using a buffer of 100 samples results in comparable performance to generative distillation in terms of FID, but not in terms of KLD. Moreover, somewhat surprisingly, increasing the buffer size from 100 to 1000 samples does not improve the FID or KLD of experience replay on CIFAR-10.

In \Cref{tab:results_cl_er_clf}, we further compare generative distillation and experience replay through the performance of a classifier that is continually trained with generative replay produced by the continually trained diffusion models (see \Cref{sec:classifier}). On Fashion MNIST, a diffusion model continually trained with experience replay using a buffer size of 100 samples already provides quite good samples to train a classifier. However, on the more complex dataset CIFAR-10, the diffusion model trained with experience replay fails to generate samples that can train a classifier well even with a buffer size of 1000 samples. 

\begin{table}[t]
    \centering    
    \caption{\label{tab:results_cl_er}\textbf{Quantitative comparison of generative distillation and experience replay.} The FID and KLD are computed after training on all tasks. Reported for each metric is mean ± standard error over 3 random seeds. The buffer size indicates the total number of samples from previous tasks that is allowed to be kept in the memory buffer.}
    \begin{tabular}{lllllll}
\toprule
& & \multicolumn{2}{c}{\bf Fashion MNIST} & & \multicolumn{2}{c}{\bf CIFAR-10} \\
Strategy & Buffer size & ~~~~$\downarrow \text{FID}$ & ~~$\downarrow \text{KLD}$ & & ~~~$\downarrow \text{FID}$ & ~~$\downarrow \text{KLD}$ \\
\midrule
    Joint \scriptsize{(upper target)} & - & 15.5 \scriptsize{$\pm$ 1.3} & 0.08 \scriptsize{$\pm$ 0.02} & & 31.9 \scriptsize{$\pm$ 2.9} & 0.12 \scriptsize{$\pm$ 0.02}\\
    Finetune \scriptsize{(lower target)} & - & 68.8 \scriptsize{$\pm$ 7.0} & 4.82 \scriptsize{$\pm$ 1.42} & & 55.8 \scriptsize{$\pm$ 2.6} & 1.14 \scriptsize{$\pm$ 0.10}\\
    \midrule
    Experience replay & 10 & 56.2 \scriptsize{$\pm$ 1.1} & 0.72 \scriptsize{$\pm$ 0.09} & & 62.0 \scriptsize{$\pm$ 4.9} & 0.48 \scriptsize{$\pm$ 0.05} \\
    Experience replay & 100 & 33.3 \scriptsize{$\pm$ 2.9} & 0.33 \scriptsize{$\pm$ 0.04} & & 42.3 \scriptsize{$\pm$ 0.9} & 0.25 \scriptsize{$\pm$ 0.03} \\
    Experience replay & 1000 & 21.6 \scriptsize{$\pm$ 0.9} & 0.28 \scriptsize{$\pm$ 0.01} & & 49.6 \scriptsize{$\pm$ 5.4} & 0.48 \scriptsize{$\pm$ 0.05} \\
    \midrule
    Generative distillation & - & 23.5 \scriptsize{$\pm$ 1.1} & 0.25 \scriptsize{$\pm$ 0.10} & & 42.3 \scriptsize{$\pm$ 4.7} & 0.15 \scriptsize{$\pm$ 0.01} \\
    \bottomrule  
\end{tabular}
\end{table}

\begin{table}[t]
    \centering
    \caption{\textbf{Comparison of generative distillation and experience replay with a continually trained classifier.} The classifier is trained on the data of the current task along with replay samples from the continually trained diffusion model. Reported is mean ± standard error over 3~seeds. The buffer size indicates the total number of samples from previous tasks that is allowed to be kept in the memory buffer for the training of the diffusion model.}\label{tab:results_cl_er_clf}
    \begin{tabular}{llclc}
        \toprule
        & & \textbf{Fashion MNIST} & & \textbf{CIFAR-10} \\
        Strategy & Buffer size & $\uparrow \text{ACC}$ & & $\uparrow \text{ACC}$ \\
        \midrule
        Joint \it \scriptsize{(upper target)} & - & 0.92\scriptsize{ $\pm$ 0.00} & & 0.77\scriptsize{ $\pm$ 0.03} \\
        Finetune \it \scriptsize{(lower target)} & - & 0.17\scriptsize{ $\pm$ 0.03} & & 0.19\scriptsize{ $\pm$ 0.00} \\
        \midrule
        Experience replay & 10 & 0.60 \scriptsize{$\pm$ 0.02} & & 0.12 \scriptsize{$\pm$ 0.01} \\
        Experience replay & 100 & 0.79 \scriptsize{$\pm$ 0.01} & & 0.12 \scriptsize{$\pm$ 0.01} \\
        Experience replay & 1000 & 0.76 \scriptsize{$\pm$ 0.03} & & 0.13 \scriptsize{$\pm$ 0.01} \\
        \midrule
        Generative distillation & - & 0.72\scriptsize{ $\pm$ 0.03} & & 0.31\scriptsize{ $\pm$ 0.03} \\
        \bottomrule
    \end{tabular}
\end{table}

\subsection{Additional quantitative results}\label{appendix:extra_experiments_results}

\begin{table}[htbp]
    \centering
    \small
    \caption{\textbf{Full comparison results among different approaches on Fashion-MNIST.} Reported for each metric is the mean ± standard error over 3 random seeds. The number of generation steps used by the generator from the previous task to train the current task model (if any) is indicated in the \textit{Teacher steps} column. All models used 10 DDIM steps to generate images for the classifier during generative replay.}
\begin{tabular}{llllll}
    \toprule
    Strategy & Teacher steps & $\lambda$ & $\downarrow \text{FID}$ & $\downarrow \text{KLD}$ & $\uparrow \text{ACC}$ \\
    \midrule
    Joint \scriptsize{(classifier)} & -  & - & ~~- & - & 0.92 \scriptsize{$\pm$ 0.00} \\
    Finetune \scriptsize{(classifier)} & -  & - & ~~- & - & 0.17 \scriptsize{$\pm$ 0.03} \\
    Joint \scriptsize{(generator)} & -  & - & ~~15.49 \scriptsize{$\pm$ 1.30} & 0.08 \scriptsize{$\pm$ 0.02} & - \\
    Finetune \scriptsize{(generator)} & -  & - & ~~68.76 \scriptsize{$\pm$ 7.00} & 4.82 \scriptsize{$\pm$ 1.42} & - \\
    \midrule
    Distil. w/ Gaussian noise & - & 8 & ~~38.87 \scriptsize{$\pm$ 2.23} & 0.85 \scriptsize{$\pm$ 0.05} & 0.63 \scriptsize{$\pm$ 0.04} \\
    Distil. w/ current data & - & 1.25 & ~~65.01 \scriptsize{$\pm$ 3.31} & 0.37 \scriptsize{$\pm$ 0.09} & 0.35 \scriptsize{$\pm$ 0.01} \\
    \midrule
    Generative replay & 2 & - & 211.79 \scriptsize{$\pm$ 6.92} & 3.39 \scriptsize{$\pm$ 0.30} & 0.50 \scriptsize{$\pm$ 0.04} \\
    Generative distillation & 2 & 0.75 & ~~\textbf{23.53 \scriptsize{$\pm$ 1.06}} & \textbf{0.25 \scriptsize{$\pm$ 0.10}} & \textbf{0.72 \scriptsize{$\pm$ 0.03}} \\
    \midrule
    Generative replay & 10 & - & 121.42 \scriptsize{$\pm$ 13.04} & 0.98 \scriptsize{$\pm$ 0.19} & 0.58 \scriptsize{$\pm$ 0.04} \\
    Generative distillation & 10 & 0.75 & ~~\textbf{20.29 \scriptsize{$\pm$ 3.07}} & \textbf{0.13 \scriptsize{$\pm$ 0.04}} & \textbf{0.79 \scriptsize{$\pm$ 0.01}} \\
    \midrule
    Generative replay & 100 & - & ~~57.54 \scriptsize{$\pm$ 8.61} & 0.61 \scriptsize{$\pm$ 0.21} & 0.66 \scriptsize{$\pm$ 0.02} \\
    Generative distillation & 100 & 0.75 & ~~\textbf{23.69 \scriptsize{$\pm$ 3.42}} & \textbf{0.12 \scriptsize{$\pm$ 0.04}} & \textbf{0.80 \scriptsize{$\pm$ 0.01}} \\
    \bottomrule
    \end{tabular}
    \label{tab:results_cl_full}
\end{table}

\begin{table}[htbp]
    \centering
    \small
    \caption{\textbf{Full comparison results among different approaches on CIFAR-10.} Reported for each metric is the mean ± standard error over 3 random seeds. The number of generation steps used by the generator from the previous task to train the current task model (if any) is indicated in the \textit{Teacher steps} column. All models used 10 DDIM steps to generate images for the classifier during generative replay.}
    \begin{tabular}{llllll}
\toprule
Strategy & Teacher steps & $\lambda$ & $\downarrow \text{FID}$ & $\downarrow \text{KLD}$ & $\uparrow \text{ACC}$ \\
\midrule
Joint \scriptsize{(classifier)} & - & - & - & - & 0.77 \scriptsize{$\pm$ 0.00}  \\
Finetune \scriptsize{(classifier)} & - & - & - & - & 0.19 \scriptsize{$\pm$ 0.00} \\
Joint \scriptsize{(generator)} & - & - & ~~31.88 \scriptsize{$\pm$ 2.89} & 0.12 \scriptsize{$\pm$ 0.02} & - \\
Finetune \scriptsize{(generator)} & - & - & ~~55.77 \scriptsize{$\pm$ 2.56} & 1.14 \scriptsize{$\pm$ 0.10} & - \\
\midrule
Distil. w/ Gaussian noise & - & 8 & ~~51.35 \scriptsize{$\pm$ 2.95} & 0.75 \scriptsize{$\pm$ 0.05} & 0.30 \scriptsize{$\pm$ 0.00} \\
Distil. w/ current data & - & 1.25 & ~~62.09 \scriptsize{$\pm$ 2.49} & 0.49 \scriptsize{$\pm$ 0.09} & 0.23 \scriptsize{$\pm$ 0.02} \\

\midrule
Generative replay & 2 & - & 166.69 \scriptsize{$\pm$ 15.95} & 1.45 \scriptsize{$\pm$ 0.45} & 0.27 \scriptsize{$\pm$ 0.02} \\
Generative distillation & 2 & 0.75 & ~~\textbf{48.82 \scriptsize{$\pm$ 2.51}} & 0.69 \scriptsize{$\pm$ 0.02} & \textbf{0.33 \scriptsize{$\pm$ 0.02}} \\
\midrule
Generative replay & 10 & - & ~~96.39 \scriptsize{$\pm$ 5.25} & 0.74 \scriptsize{$\pm$ 0.14} & 0.26 \scriptsize{$\pm$ 0.01} \\
Generative distillation & 10 & 0.75 & ~~\textbf{42.34 \scriptsize{$\pm$ 4.74}} & \textbf{0.15 \scriptsize{$\pm$ 0.01}} & \textbf{0.31 \scriptsize{$\pm$ 0.03}} \\
\midrule
Generative replay & 100 & - & ~~66.47 \scriptsize{$\pm$ 7.39} & 0.43 \scriptsize{$\pm$ 0.07} & 0.28 \scriptsize{$\pm$ 0.02} \\
Generative distillation & 100 & 0.75 & ~~\textbf{30.53 \scriptsize{$\pm$ 1.29}} & 0.27 \scriptsize{$\pm$ 0.05} & \textbf{0.35 \scriptsize{$\pm$ 0.03}} \\
\bottomrule
\end{tabular}
    \label{tab:results_cl_full_cifar}
\end{table}

\begin{figure}[ht]
    \centering

    \begin{subfigure}{\textwidth}
    \begin{subfigure}[b]{0.471\textwidth}
         \centering
         \includegraphics{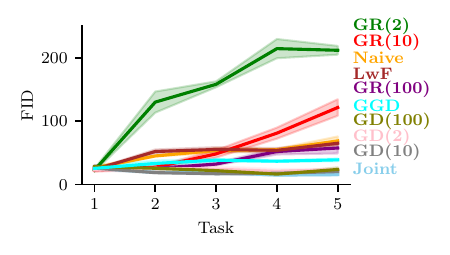}
    \end{subfigure}
    \hfill
    \begin{subfigure}[b]{0.471\textwidth}
         \centering
         \includegraphics{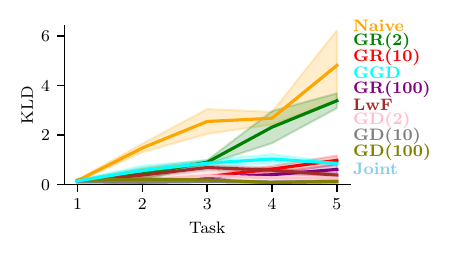}
    \end{subfigure}
    \vspace{-0.1in}
    \caption{Fashion-MNIST}
    \end{subfigure}
    
    \begin{subfigure}{\textwidth}
    \begin{subfigure}[b]{0.471\textwidth}
        \centering
        \includegraphics{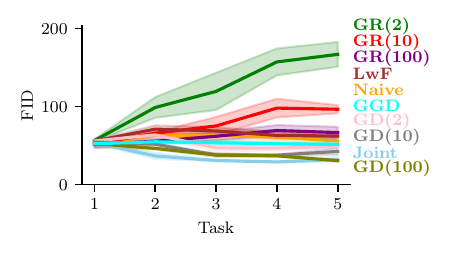}
    \end{subfigure}
    \hfill
    \begin{subfigure}[b]{0.471\textwidth}
        \centering
        \includegraphics{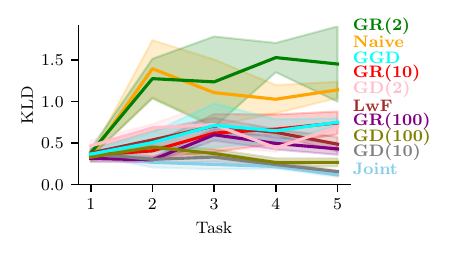}
    \end{subfigure}
    \vspace{-0.1in}
    \caption{CIFAR-10}
    \end{subfigure}
         
    \caption{\textbf{Quality of the diffusion model throughout the continual learning process (with additional approaches).} Each point represents a metric achieved after training in a specific task. Displayed are the means ± standard errors over 3 random seeds. Joint: training using data from all tasks seen so far (upper target), Naive: fine-tuning on the current task (lower target), GR($n$): generative replay with a teacher using $n$ DDIM steps, LwF: distillation with current data, GGD: distillation with Gaussian noise, GD: generative distillation.}
    \label{fig:app_results_cl_metrics_through_tasks}
\end{figure}

To report the full results of the additional experiments described in \Cref{experiments:extra}, we provide \Cref{tab:results_cl_full,tab:results_cl_full_cifar} (quantitative results) and \Cref{fig:app_results_cl_metrics_through_tasks} (graph of quality metrics throughout the continual learning process).

\clearpage
\subsection{Samples throughout the continual learning process}

In this section, we display samples generated by the diffusion model throughout the continual learning process following the different approaches we have discussed. This allows the reader to qualitatively evaluate the diffusion model's performance under each approach and throughout their continual training. The samples were obtained using a consistent representative random seed across all tasks and approaches. Note that we could not make sure that all the experiments were executed on the same hardware, thus some runs may vary slightly (refer to \hyperlink{https://pytorch.org/docs/stable/notes/randomness.html}{https://pytorch.org/docs/stable/notes/randomness.html}).

\begin{figure}[htbp]
\centering
\begin{subfigure}[b]{0.195\textwidth}
     \centering
     \includegraphics[width=\textwidth]{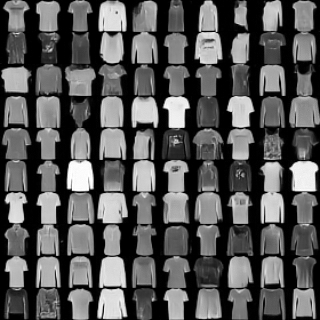}
     \caption{Task 1}
\end{subfigure}
\begin{subfigure}[b]{0.195\textwidth}
     \centering
     \includegraphics[width=\textwidth]{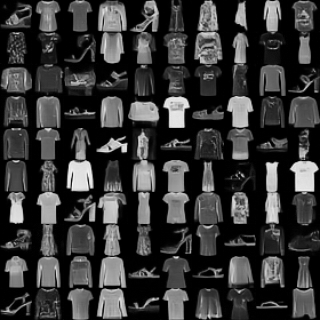}
     \caption{Task 2}
\end{subfigure}
\begin{subfigure}[b]{0.195\textwidth}
     \centering
     \includegraphics[width=\textwidth]{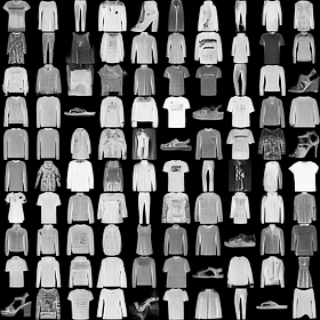}
     \caption{Task 3}
\end{subfigure}
\begin{subfigure}[b]{0.195\textwidth}
     \centering
     \includegraphics[width=\textwidth]{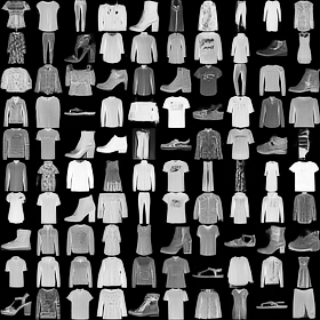}
     \caption{Task 4}
\end{subfigure}
\begin{subfigure}[b]{0.195\textwidth}
     \centering
     \includegraphics[width=\textwidth]{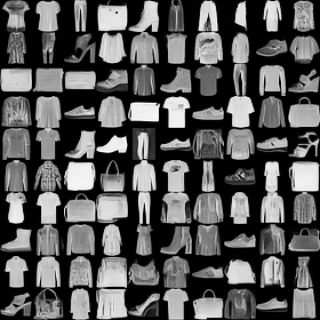}
     \caption{Task 5}
\end{subfigure}
\caption{\textbf{Samples for joint training on Fashion-MNIST.} This can be interpreted as an upper target.\label{fig:joint}}
\end{figure}

\begin{figure}[htbp]
\centering
\begin{subfigure}[b]{0.195\textwidth}
     \centering
     \includegraphics[width=\textwidth]{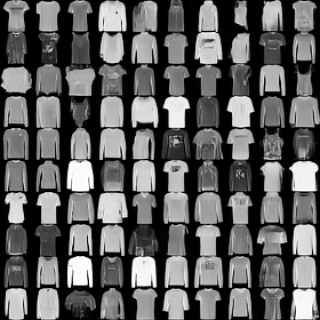}
     \caption{Task 1}
\end{subfigure}
\begin{subfigure}[b]{0.195\textwidth}
     \centering
     \includegraphics[width=\textwidth]{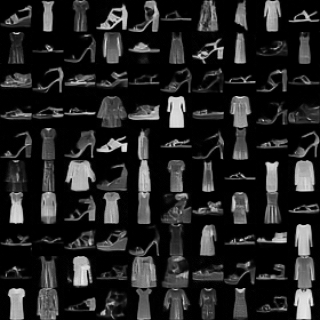}
     \caption{Task 2}
\end{subfigure}
\begin{subfigure}[b]{0.195\textwidth}
     \centering
     \includegraphics[width=\textwidth]{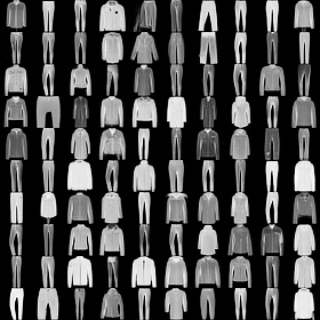}
     \caption{Task 3}
\end{subfigure}
\begin{subfigure}[b]{0.195\textwidth}
     \centering
     \includegraphics[width=\textwidth]{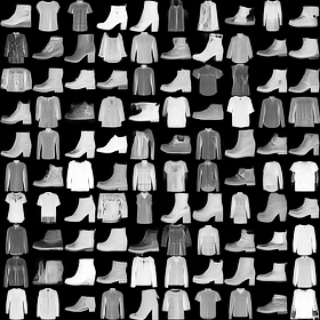}
     \caption{Task 4}
\end{subfigure}
\begin{subfigure}[b]{0.195\textwidth}
     \centering
     \includegraphics[width=\textwidth]{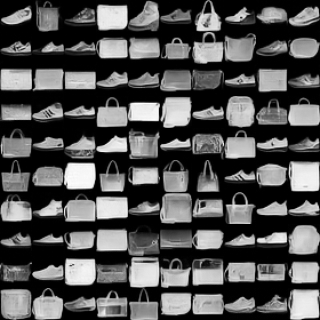}
     \caption{Task 5}
\end{subfigure}
\caption{\textbf{Samples for finetune on Fashion-MNIST.} This can be interpreted as a lower target.}
\end{figure}

\begin{figure}[htbp]
\centering
\begin{subfigure}[b]{0.195\textwidth}
     \centering
     \includegraphics[width=\textwidth]{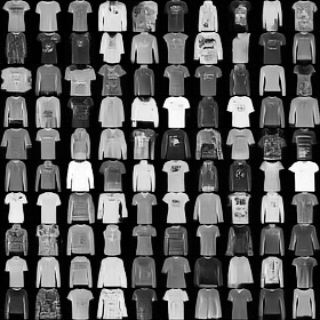}
     \caption{Task 1}
\end{subfigure}
\begin{subfigure}[b]{0.195\textwidth}
     \centering
     \includegraphics[width=\textwidth]{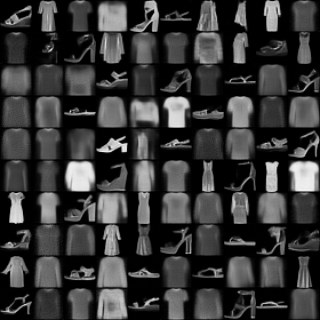}
     \caption{Task 2}
\end{subfigure}
\begin{subfigure}[b]{0.195\textwidth}
     \centering
     \includegraphics[width=\textwidth]{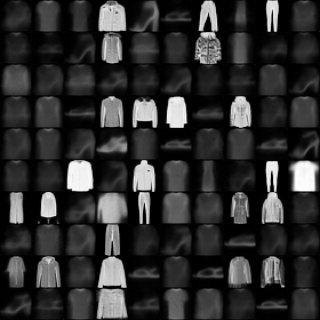}
     \caption{Task 3}
\end{subfigure}
\begin{subfigure}[b]{0.195\textwidth}
     \centering
     \includegraphics[width=\textwidth]{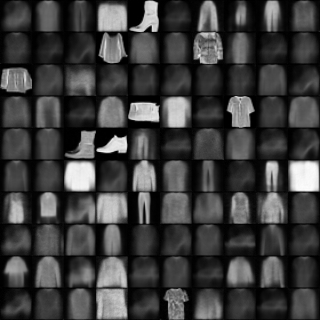}
     \caption{Task 4}
\end{subfigure}
\begin{subfigure}[b]{0.195\textwidth}
     \centering
     \includegraphics[width=\textwidth]{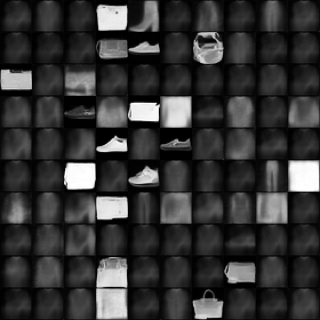}
     \caption{Task 5}
\end{subfigure}
\caption{\textbf{Samples for generative replay (2 teacher DDIM steps) on Fashion-MNIST.} Standard generative replay causes a catastrophic degradation of the denoising capabilities of the diffusion model for samples from previous tasks.}
\label{fig:qualitative_gr2}
\end{figure}

\begin{figure}[htbp]
\centering
\begin{subfigure}[b]{0.195\textwidth}
     \centering
     \includegraphics[width=\textwidth]{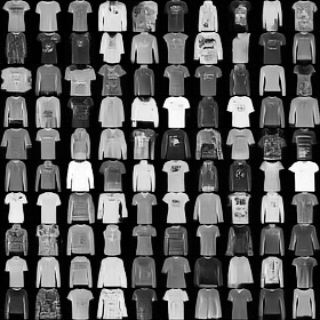}
     \caption{Task 1}
\end{subfigure}
\begin{subfigure}[b]{0.195\textwidth}
     \centering
     \includegraphics[width=\textwidth]{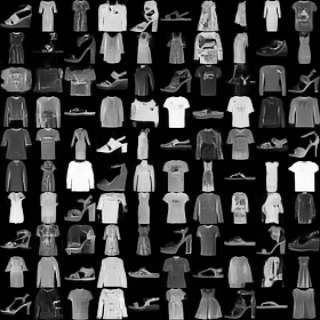}
     \caption{Task 2}
\end{subfigure}
\begin{subfigure}[b]{0.195\textwidth}
     \centering
     \includegraphics[width=\textwidth]{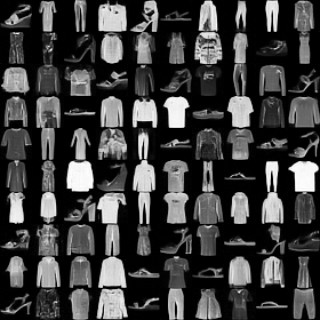}
     \caption{Task 3}
\end{subfigure}
\begin{subfigure}[b]{0.195\textwidth}
     \centering
     \includegraphics[width=\textwidth]{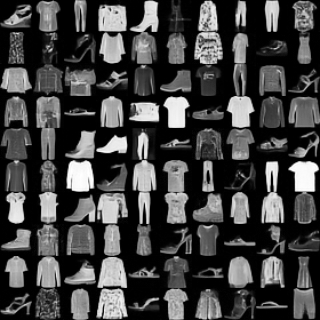}
     \caption{Task 4}
\end{subfigure}
\begin{subfigure}[b]{0.195\textwidth}
     \centering
     \includegraphics[width=\textwidth]{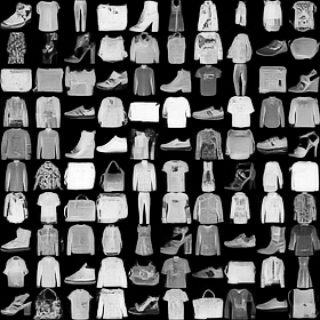}
     \caption{Task 5}
\end{subfigure}
\caption{\textbf{Samples for generative distillation (2 teacher DDIM steps) on Fashion-MNIST.} Generative distillation markedly improves the quality with only a moderate increase in the computational cost.}
\label{fig:qualitative_gd2}
\end{figure}

\begin{figure}[htbp]
\centering
\begin{subfigure}[b]{0.195\textwidth}
     \centering
     \includegraphics[width=\textwidth]{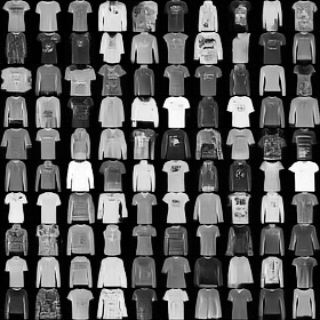}
     \caption{Task 1}
\end{subfigure}
\begin{subfigure}[b]{0.195\textwidth}
     \centering
     \includegraphics[width=\textwidth]{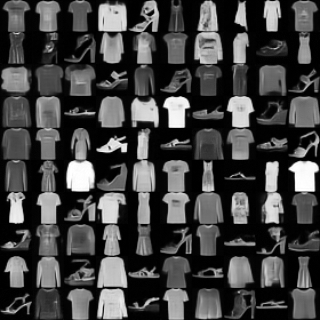}
     \caption{Task 2}
\end{subfigure}
\begin{subfigure}[b]{0.195\textwidth}
     \centering
     \includegraphics[width=\textwidth]{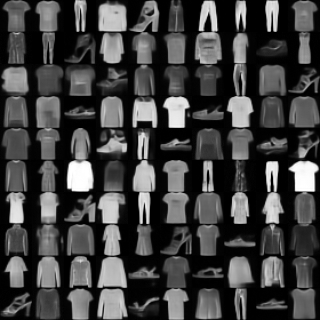}
     \caption{Task 3}
\end{subfigure}
\begin{subfigure}[b]{0.195\textwidth}
     \centering
     \includegraphics[width=\textwidth]{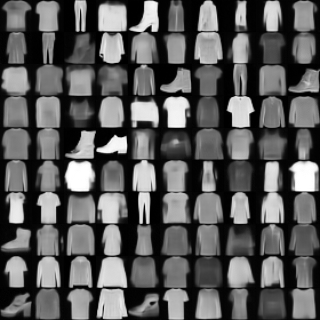}
     \caption{Task 4}
\end{subfigure}
\begin{subfigure}[b]{0.195\textwidth}
     \centering
     \includegraphics[width=\textwidth]{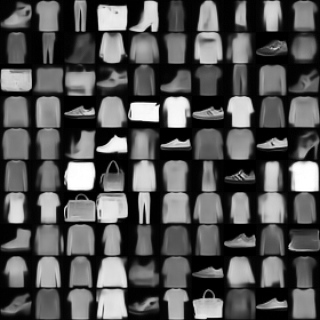}
     \caption{Task 5}
\end{subfigure}
\caption{\textbf{Samples for generative replay (10 teacher DDIM steps) on Fashion-MNIST.} Increasing the number of teacher DDIM steps reduces but does not solve the catastrophic loss in denoising capabilities.}
\label{fig:qualitative_gr10}
\end{figure}

\begin{figure}[htbp]
\centering
\begin{subfigure}[b]{0.195\textwidth}
     \centering
     \includegraphics[width=\textwidth]{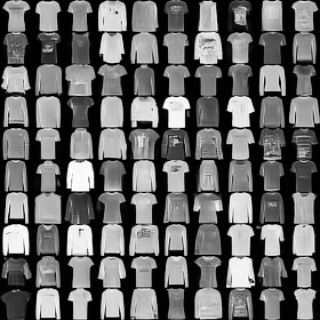}
     \caption{Task 1}
\end{subfigure}
\begin{subfigure}[b]{0.195\textwidth}
     \centering
     \includegraphics[width=\textwidth]{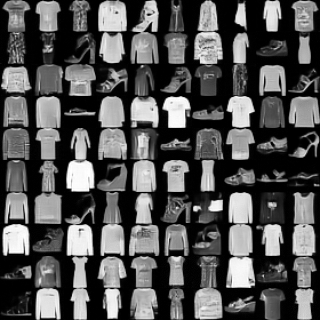}
     \caption{Task 2}
\end{subfigure}
\begin{subfigure}[b]{0.195\textwidth}
     \centering
     \includegraphics[width=\textwidth]{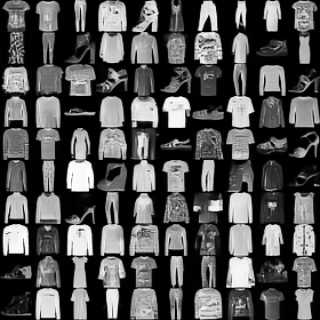}
     \caption{Task 3}
\end{subfigure}
\begin{subfigure}[b]{0.195\textwidth}
     \centering
     \includegraphics[width=\textwidth]{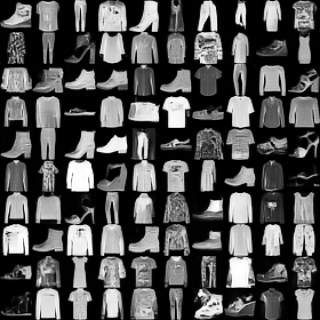}
     \caption{Task 4}
\end{subfigure}
\begin{subfigure}[b]{0.195\textwidth}
     \centering
     \includegraphics[width=\textwidth]{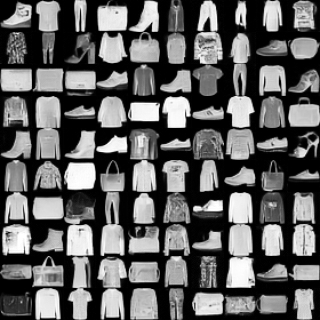}
     \caption{Task 5}
\end{subfigure}
\caption{\textbf{Samples for generative distillation (10 teacher DDIM steps) on Fashion-MNIST.} Generative distillation markedly improves the quality with only a moderate increase in the computational cost.}
\label{fig:qualitative_gd10}
\end{figure}

\begin{figure}[htbp]
\centering
\begin{subfigure}[b]{0.195\textwidth}
     \centering
     \includegraphics[width=\textwidth]{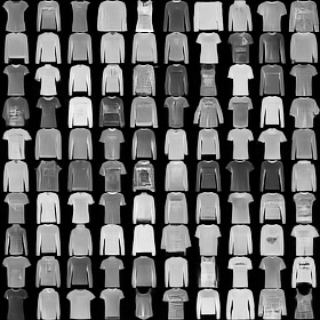}
     \caption{Task 1}
\end{subfigure}
\begin{subfigure}[b]{0.195\textwidth}
     \centering
     \includegraphics[width=\textwidth]{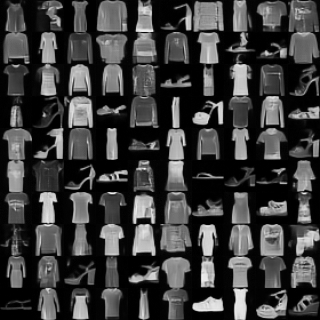}
     \caption{Task 2}
\end{subfigure}
\begin{subfigure}[b]{0.195\textwidth}
     \centering
     \includegraphics[width=\textwidth]{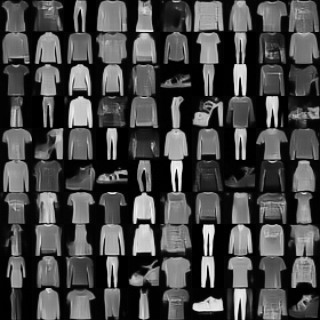}
     \caption{Task 3}
\end{subfigure}
\begin{subfigure}[b]{0.195\textwidth}
     \centering
     \includegraphics[width=\textwidth]{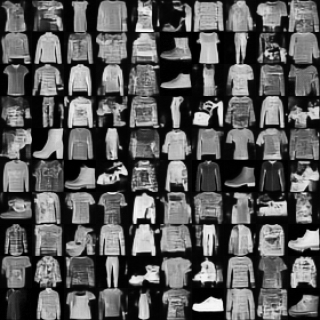}
     \caption{Task 4}
\end{subfigure}
\begin{subfigure}[b]{0.195\textwidth}
     \centering
     \includegraphics[width=\textwidth]{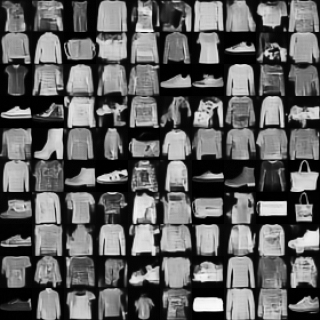}
     \caption{Task 5}
\end{subfigure}
\caption{\textbf{Samples for generative replay (100 teacher DDIM steps) on Fashion-MNIST.} Increasing the number of teacher DDIM steps reduces but does not solve the catastrophic loss in denoising capabilities.}
\label{fig:qualitative_gr100}
\end{figure}

\begin{figure}[htbp]
\centering
\begin{subfigure}[b]{0.195\textwidth}
     \centering
     \includegraphics[width=\textwidth]{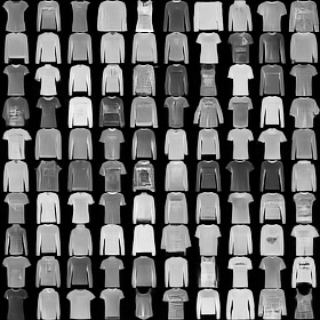}
     \caption{Task 1}
\end{subfigure}
\begin{subfigure}[b]{0.195\textwidth}
     \centering
     \includegraphics[width=\textwidth]{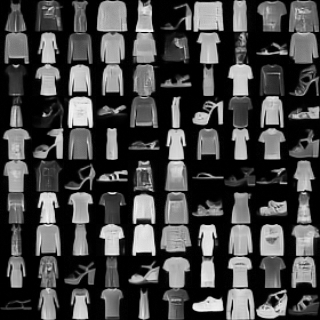}
     \caption{Task 2}
\end{subfigure}
\begin{subfigure}[b]{0.195\textwidth}
     \centering
     \includegraphics[width=\textwidth]{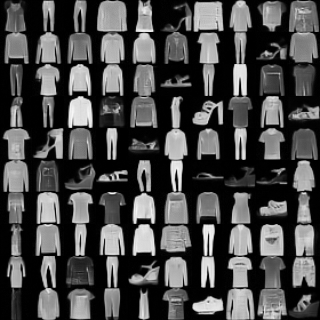}
     \caption{Task 3}
\end{subfigure}
\begin{subfigure}[b]{0.195\textwidth}
     \centering
     \includegraphics[width=\textwidth]{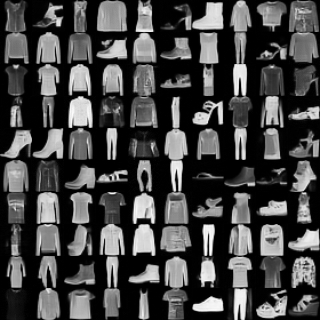}
     \caption{Task 4}
\end{subfigure}
\begin{subfigure}[b]{0.195\textwidth}
     \centering
     \includegraphics[width=\textwidth]{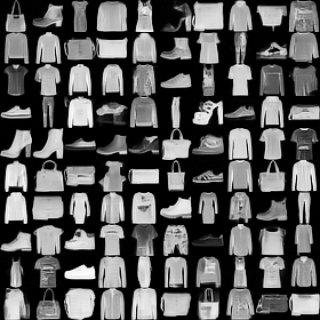}
     \caption{Task 5}
\end{subfigure}
\caption{\textbf{Samples for generative distillation (100 teacher DDIM steps) on Fashion-MNIST.} Generative distillation markedly improves the quality with only a moderate increase in the computational cost.}
\label{fig:qualitative_gd100}
\end{figure}

\begin{figure}[htbp]
\centering
\begin{subfigure}[b]{0.195\textwidth}
     \centering
     \includegraphics[width=\textwidth]{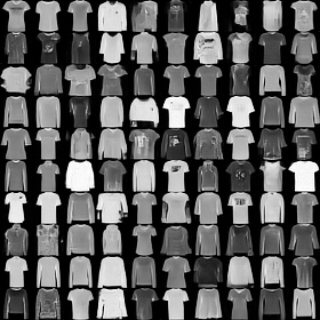}
     \caption{Task 1}
\end{subfigure}
\begin{subfigure}[b]{0.195\textwidth}
     \centering
     \includegraphics[width=\textwidth]{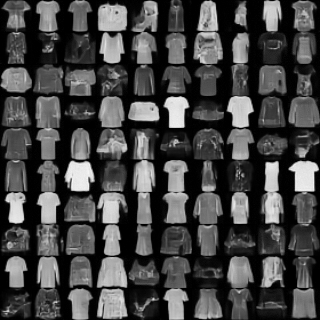}
     \caption{Task 2}
\end{subfigure}
\begin{subfigure}[b]{0.195\textwidth}
     \centering
     \includegraphics[width=\textwidth]{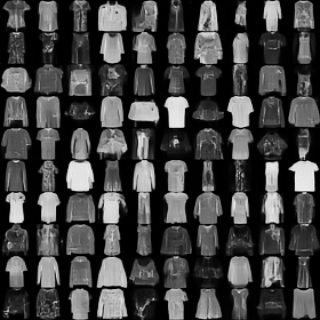}
     \caption{Task 3}
\end{subfigure}
\begin{subfigure}[b]{0.195\textwidth}
     \centering
     \includegraphics[width=\textwidth]{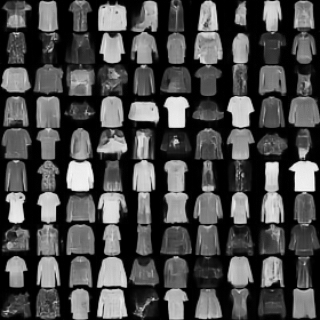}
     \caption{Task 4}
\end{subfigure}
\begin{subfigure}[b]{0.195\textwidth}
     \centering
     \includegraphics[width=\textwidth]{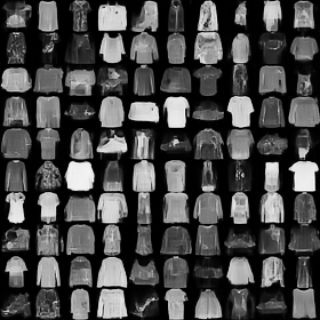}
     \caption{Task 5}
\end{subfigure}
\caption{\textbf{Samples for LwF-like distillation on Fashion-MNIST.} The LwF-like approach improves over naïve generative replay but still has an image quality degradation over the continual learning process.}
\end{figure}

\begin{figure}[htbp]
\centering
\begin{subfigure}[b]{0.195\textwidth}
     \centering
     \includegraphics[width=\textwidth]{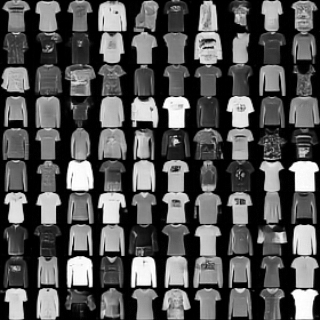}
     \caption{Task 1}
\end{subfigure}
\begin{subfigure}[b]{0.195\textwidth}
     \centering
     \includegraphics[width=\textwidth]{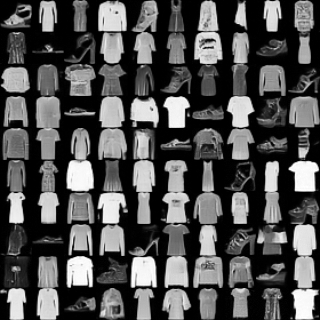}
     \caption{Task 2}
\end{subfigure}
\begin{subfigure}[b]{0.195\textwidth}
     \centering
     \includegraphics[width=\textwidth]{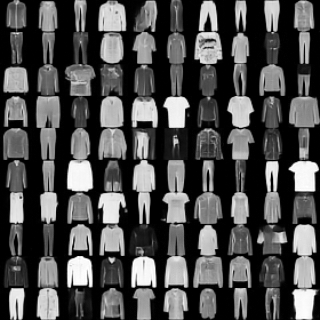}
     \caption{Task 3}
\end{subfigure}
\begin{subfigure}[b]{0.195\textwidth}
     \centering
     \includegraphics[width=\textwidth]{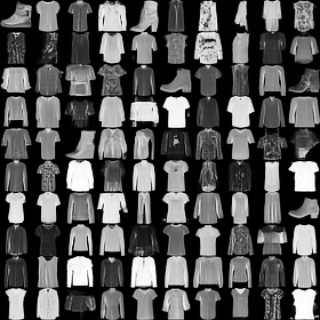}
     \caption{Task 4}
\end{subfigure}
\begin{subfigure}[b]{0.195\textwidth}
     \centering
     \includegraphics[width=\textwidth]{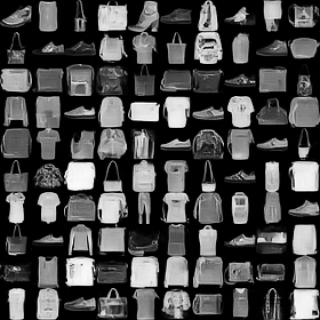}
     \caption{Task 5}
\end{subfigure}
\caption{\textbf{Samples for distillation with Gaussian noise on Fashion-MNIST.} Similar to distilling with current data, distilling with noise preserves knowledge of classes similar to the ones in the last task learned.\label{fig:gaussian_noise}}
\label{fig:qualitative_gaussian}
\end{figure}


\begin{figure}[htbp]
\centering
\begin{subfigure}[b]{0.195\textwidth}
     \centering
     \includegraphics[width=\textwidth]{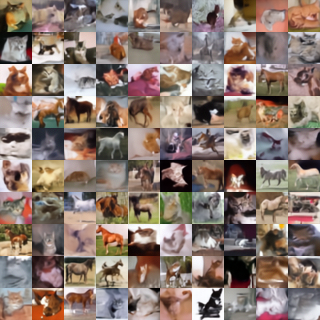}
     \caption{Task 1}
\end{subfigure}
\begin{subfigure}[b]{0.195\textwidth}
     \centering
     \includegraphics[width=\textwidth]{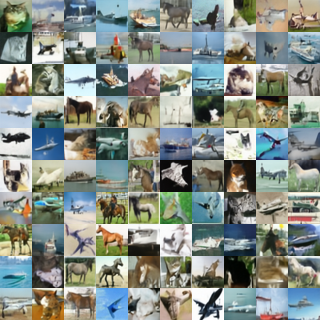}
     \caption{Task 2}
\end{subfigure}
\begin{subfigure}[b]{0.195\textwidth}
     \centering
     \includegraphics[width=\textwidth]{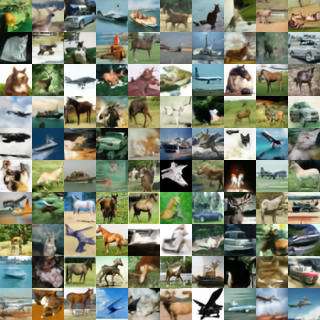}
     \caption{Task 3}
\end{subfigure}
\begin{subfigure}[b]{0.195\textwidth}
     \centering
     \includegraphics[width=\textwidth]{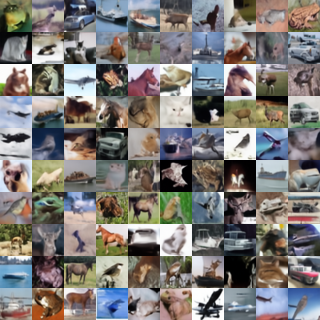}
     \caption{Task 4}
\end{subfigure}
\begin{subfigure}[b]{0.195\textwidth}
     \centering
     \includegraphics[width=\textwidth]{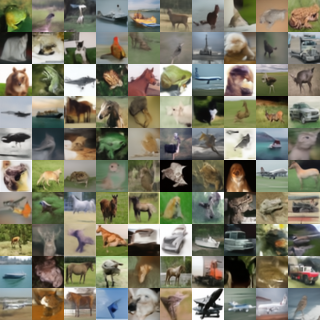}
     \caption{Task 5}
\end{subfigure}
\caption{\textbf{Samples for joint training on CIFAR-10.} This can be interpreted as an upper target.\label{fig:joint_cifar}}
\end{figure}

\begin{figure}[htbp]
\centering
\begin{subfigure}[b]{0.195\textwidth}
     \centering
     \includegraphics[width=\textwidth]{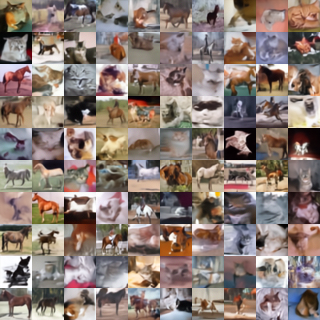}
     \caption{Task 1}
\end{subfigure}
\begin{subfigure}[b]{0.195\textwidth}
     \centering
     \includegraphics[width=\textwidth]{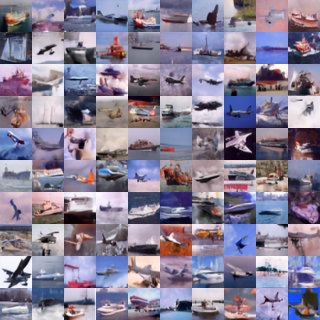}
     \caption{Task 2}
\end{subfigure}
\begin{subfigure}[b]{0.195\textwidth}
     \centering
     \includegraphics[width=\textwidth]{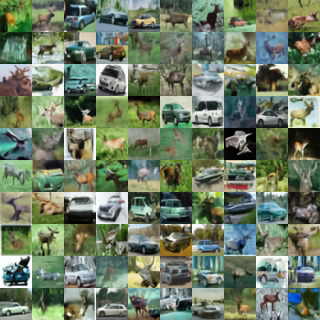}
     \caption{Task 3}
\end{subfigure}
\begin{subfigure}[b]{0.195\textwidth}
     \centering
     \includegraphics[width=\textwidth]{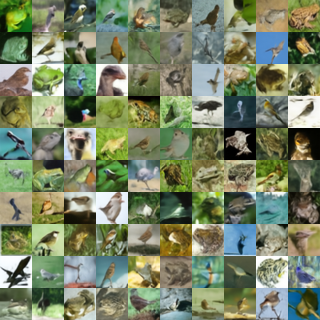}
     \caption{Task 4}
\end{subfigure}
\begin{subfigure}[b]{0.195\textwidth}
     \centering
     \includegraphics[width=\textwidth]{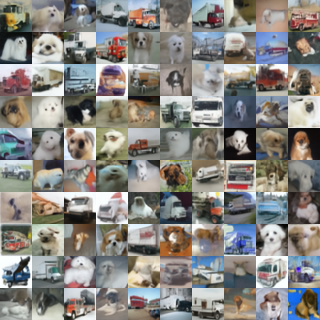}
     \caption{Task 5}
\end{subfigure}
\caption{\textbf{Samples for finetune on CIFAR-10.} This can be interpreted as a lower target.}
\end{figure}

\begin{figure}[htbp]
\centering
\begin{subfigure}[b]{0.195\textwidth}
     \centering
     \includegraphics[width=\textwidth]{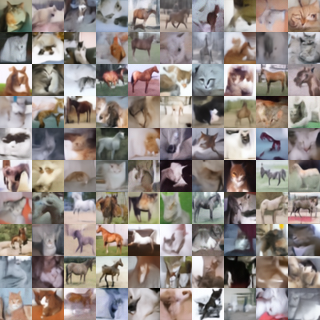}
     \caption{Task 1}
\end{subfigure}
\begin{subfigure}[b]{0.195\textwidth}
     \centering
     \includegraphics[width=\textwidth]{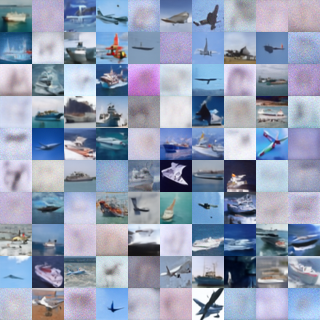}
     \caption{Task 2}
\end{subfigure}
\begin{subfigure}[b]{0.195\textwidth}
     \centering
     \includegraphics[width=\textwidth]{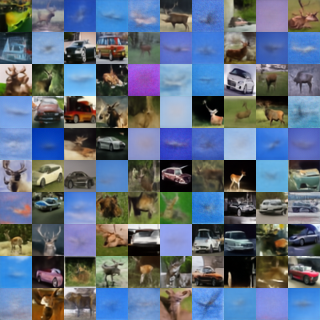}
     \caption{Task 3}
\end{subfigure}
\begin{subfigure}[b]{0.195\textwidth}
     \centering
     \includegraphics[width=\textwidth]{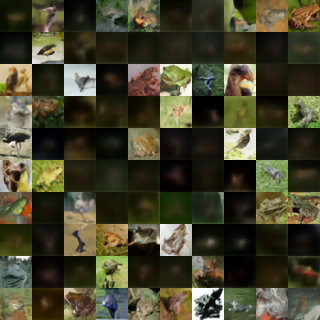}
     \caption{Task 4}
\end{subfigure}
\begin{subfigure}[b]{0.195\textwidth}
     \centering
     \includegraphics[width=\textwidth]{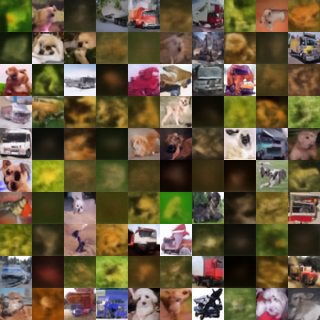}
     \caption{Task 5}
\end{subfigure}
\caption{\textbf{Samples for generative replay (2 teacher DDIM steps) on CIFAR-10.} Standard generative replay causes a catastrophic degradation of the denoising capabilities of the diffusion model for samples from previous tasks.}
\label{fig:qualitative_gr2_cifar}
\end{figure}

\begin{figure}[htbp]
\centering
\begin{subfigure}[b]{0.195\textwidth}
     \centering
     \includegraphics[width=\textwidth]{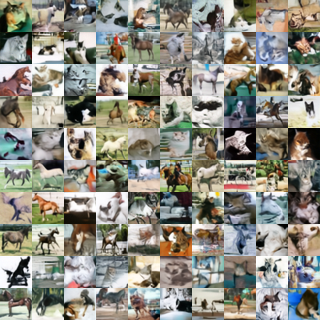}
     \caption{Task 1}
\end{subfigure}
\begin{subfigure}[b]{0.195\textwidth}
     \centering
     \includegraphics[width=\textwidth]{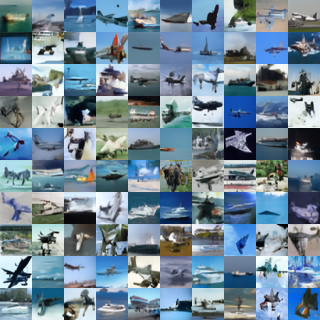}
     \caption{Task 2}
\end{subfigure}
\begin{subfigure}[b]{0.195\textwidth}
     \centering
     \includegraphics[width=\textwidth]{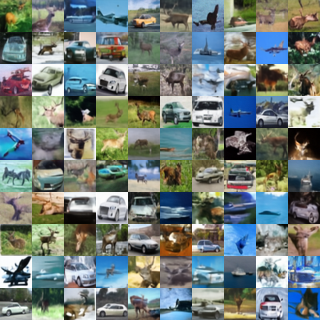}
     \caption{Task 3}
\end{subfigure}
\begin{subfigure}[b]{0.195\textwidth}
     \centering
     \includegraphics[width=\textwidth]{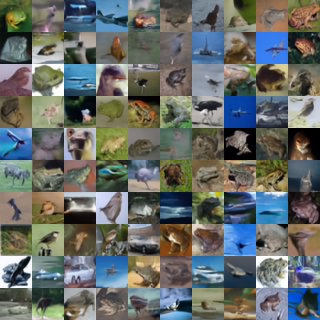}
     \caption{Task 4}
\end{subfigure}
\begin{subfigure}[b]{0.195\textwidth}
     \centering
     \includegraphics[width=\textwidth]{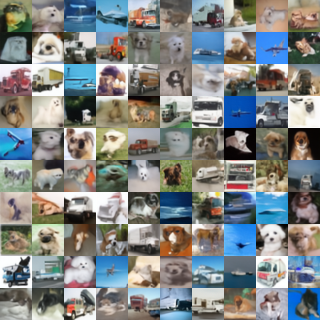}
     \caption{Task 5}
\end{subfigure}
\caption{\textbf{Samples for generative distillation (2 teacher DDIM steps) on CIFAR-10.} Generative distillation markedly improves the quality with only a moderate increase in the computational cost.}
\label{fig:qualitative_gd2_cifar}
\end{figure}

\begin{figure}[htbp]
\centering
\begin{subfigure}[b]{0.195\textwidth}
     \centering
     \includegraphics[width=\textwidth]{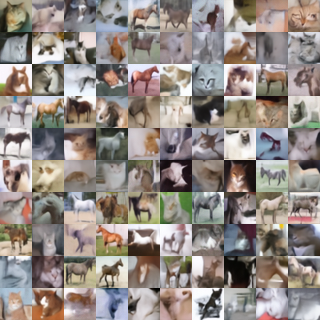}
     \caption{Task 1}
\end{subfigure}
\begin{subfigure}[b]{0.195\textwidth}
     \centering
     \includegraphics[width=\textwidth]{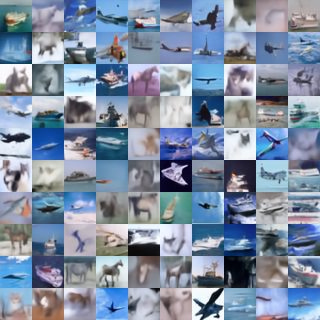}
     \caption{Task 2}
\end{subfigure}
\begin{subfigure}[b]{0.195\textwidth}
     \centering
     \includegraphics[width=\textwidth]{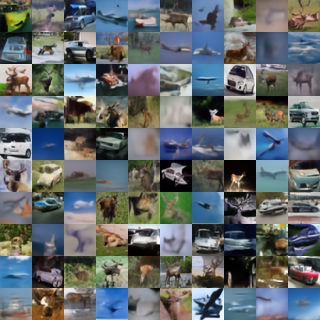}
     \caption{Task 3}
\end{subfigure}
\begin{subfigure}[b]{0.195\textwidth}
     \centering
     \includegraphics[width=\textwidth]{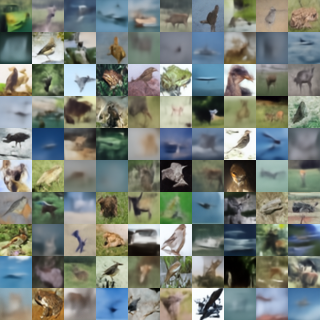}
     \caption{Task 4}
\end{subfigure}
\begin{subfigure}[b]{0.195\textwidth}
     \centering
     \includegraphics[width=\textwidth]{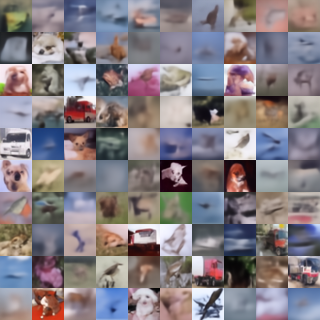}
     \caption{Task 5}
\end{subfigure}
\caption{\textbf{Samples for generative replay (10 teacher DDIM steps) on CIFAR-10.} Increasing the number of teacher DDIM steps reduces but does not solve the catastrophic loss in denoising capabilities.}
\label{fig:qualitative_gr10_cifar}
\end{figure}

\begin{figure}[htbp]
\centering
\begin{subfigure}[b]{0.195\textwidth}
     \centering
     \includegraphics[width=\textwidth]{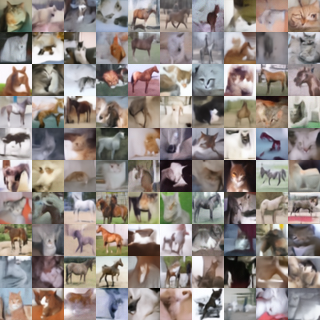}
     \caption{Task 1}
\end{subfigure}
\begin{subfigure}[b]{0.195\textwidth}
     \centering
     \includegraphics[width=\textwidth]{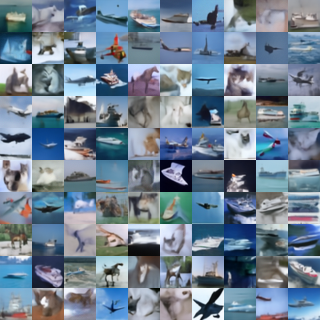}
     \caption{Task 2}
\end{subfigure}
\begin{subfigure}[b]{0.195\textwidth}
     \centering
     \includegraphics[width=\textwidth]{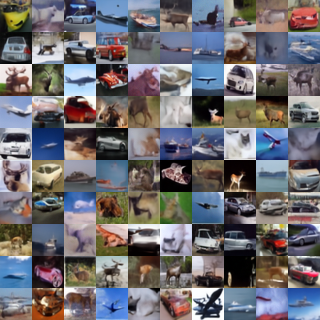}
     \caption{Task 3}
\end{subfigure}
\begin{subfigure}[b]{0.195\textwidth}
     \centering
     \includegraphics[width=\textwidth]{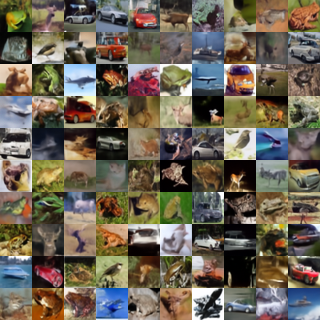}
     \caption{Task 4}
\end{subfigure}
\begin{subfigure}[b]{0.195\textwidth}
     \centering
     \includegraphics[width=\textwidth]{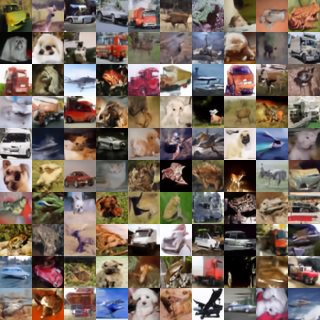}
     \caption{Task 5}
\end{subfigure}
\caption{\textbf{Samples for generative distillation (10 teacher DDIM steps) on CIFAR-10.} Generative distillation markedly improves the quality with only a moderate increase in the computational cost.}
\label{fig:qualitative_gd10_cifar}
\end{figure}

\begin{figure}[htbp]
\centering
\begin{subfigure}[b]{0.195\textwidth}
     \centering
     \includegraphics[width=\textwidth]{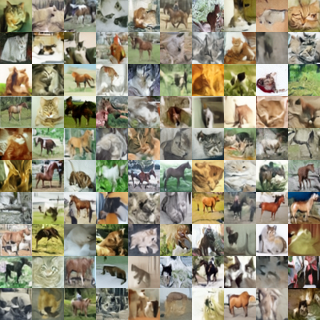}
     \caption{Task 1}
\end{subfigure}
\begin{subfigure}[b]{0.195\textwidth}
     \centering
     \includegraphics[width=\textwidth]{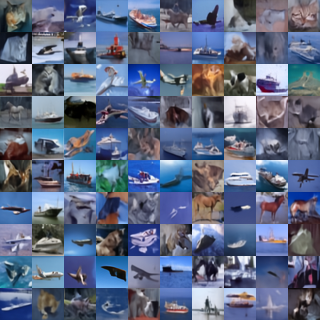}
     \caption{Task 2}
\end{subfigure}
\begin{subfigure}[b]{0.195\textwidth}
     \centering
     \includegraphics[width=\textwidth]{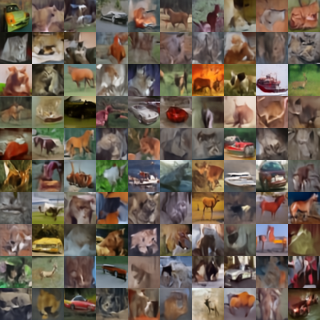}
     \caption{Task 3}
\end{subfigure}
\begin{subfigure}[b]{0.195\textwidth}
     \centering
     \includegraphics[width=\textwidth]{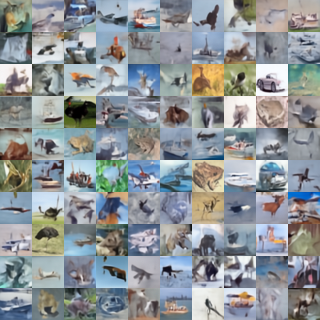}
     \caption{Task 4}
\end{subfigure}
\begin{subfigure}[b]{0.195\textwidth}
     \centering
     \includegraphics[width=\textwidth]{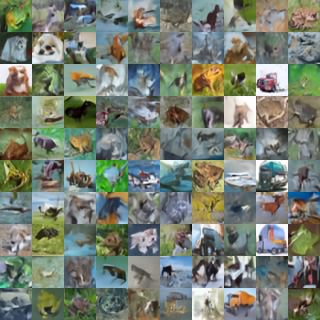}
     \caption{Task 5}
\end{subfigure}
\caption{\textbf{Samples for generative replay (100 teacher DDIM steps) on CIFAR-10.} Increasing the number of teacher DDIM steps reduces but does not solve the catastrophic loss in denoising capabilities.}
\label{fig:qualitative_gr100_cifar}
\end{figure}

\begin{figure}[htbp]
\centering
\begin{subfigure}[b]{0.195\textwidth}
     \centering
     \includegraphics[width=\textwidth]{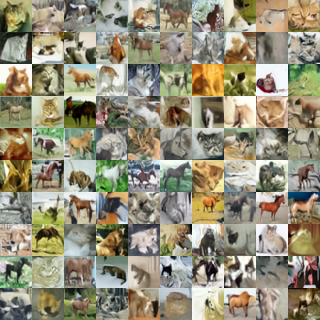}
     \caption{Task 1}
\end{subfigure}
\begin{subfigure}[b]{0.195\textwidth}
     \centering
     \includegraphics[width=\textwidth]{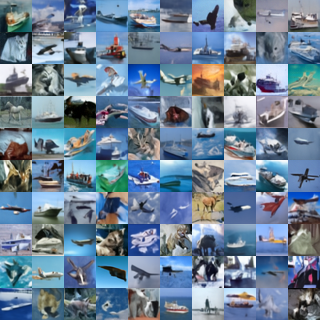}
     \caption{Task 2}
\end{subfigure}
\begin{subfigure}[b]{0.195\textwidth}
     \centering
     \includegraphics[width=\textwidth]{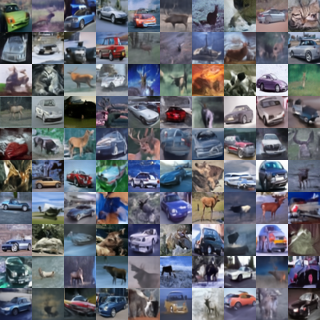}
     \caption{Task 3}
\end{subfigure}
\begin{subfigure}[b]{0.195\textwidth}
     \centering
     \includegraphics[width=\textwidth]{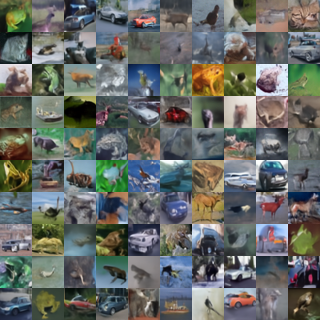}
     \caption{Task 4}
\end{subfigure}
\begin{subfigure}[b]{0.195\textwidth}
     \centering
     \includegraphics[width=\textwidth]{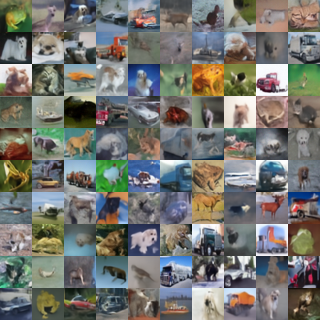}
     \caption{Task 5}
\end{subfigure}
\caption{\textbf{Samples for generative distillation (100 teacher DDIM steps) on CIFAR-10.} Generative distillation markedly improves the quality with only a moderate increase in the computational cost.}
\label{fig:qualitative_gd100_cifar}
\end{figure}

\begin{figure}[htbp]
\centering
\begin{subfigure}[b]{0.195\textwidth}
     \centering
     \includegraphics[width=\textwidth]{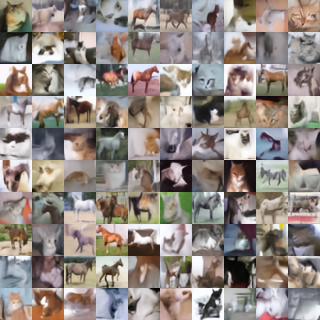}
     \caption{Task 1}
\end{subfigure}
\begin{subfigure}[b]{0.195\textwidth}
     \centering
     \includegraphics[width=\textwidth]{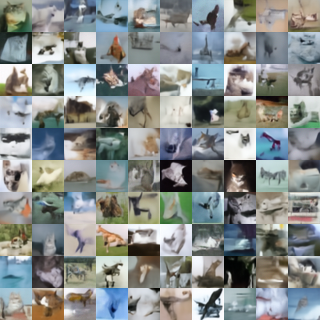}
     \caption{Task 2}
\end{subfigure}
\begin{subfigure}[b]{0.195\textwidth}
     \centering
     \includegraphics[width=\textwidth]{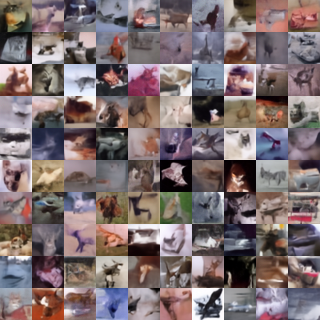}
     \caption{Task 3}
\end{subfigure}
\begin{subfigure}[b]{0.195\textwidth}
     \centering
     \includegraphics[width=\textwidth]{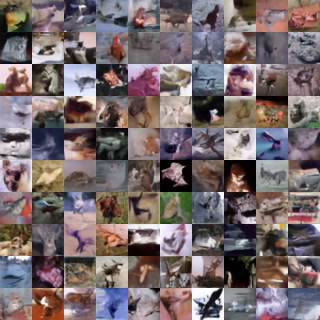}
     \caption{Task 4}
\end{subfigure}
\begin{subfigure}[b]{0.195\textwidth}
     \centering
     \includegraphics[width=\textwidth]{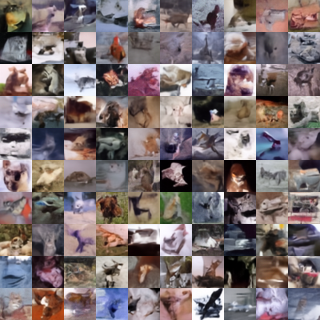}
     \caption{Task 5}
\end{subfigure}
\caption{\textbf{Samples for LwF-like distillation on CIFAR-10.} The LwF-like approach improves over naïve generative replay but still has an image quality degradation over the continual learning process.}
\end{figure}

\begin{figure}[htbp]
\centering
\begin{subfigure}[b]{0.195\textwidth}
     \centering
     \includegraphics[width=\textwidth]{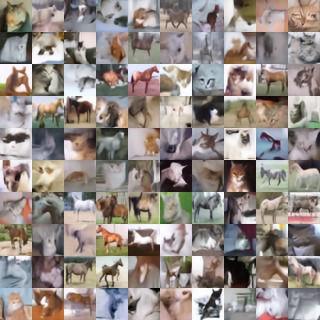}
     \caption{Task 1}
\end{subfigure}
\begin{subfigure}[b]{0.195\textwidth}
     \centering
     \includegraphics[width=\textwidth]{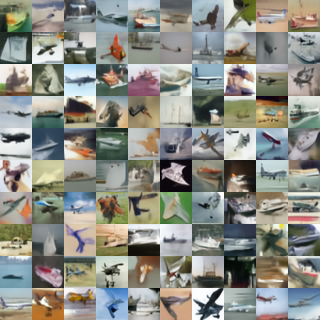}
     \caption{Task 2}
\end{subfigure}
\begin{subfigure}[b]{0.195\textwidth}
     \centering
     \includegraphics[width=\textwidth]{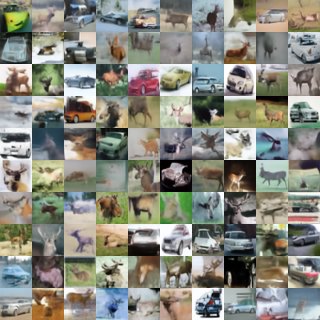}
     \caption{Task 3}
\end{subfigure}
\begin{subfigure}[b]{0.195\textwidth}
     \centering
     \includegraphics[width=\textwidth]{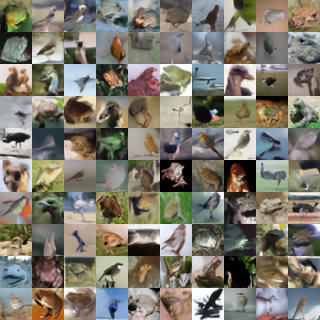}
     \caption{Task 4}
\end{subfigure}
\begin{subfigure}[b]{0.195\textwidth}
     \centering
     \includegraphics[width=\textwidth]{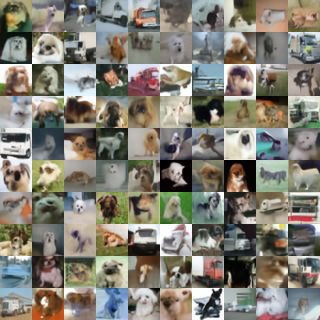}
     \caption{Task 5}
\end{subfigure}
\caption{\textbf{Samples for distillation with Gaussian noise on CIFAR-10.} Similar to distilling with current data, distilling with noise preserves knowledge of classes similar to the ones in the last task learned.}
\label{fig:qualitative_gaussian_cifar}
\end{figure}

\end{document}